\documentclass[accepted]{uai2023} %

\usepackage[american]{babel}

\usepackage{natbib} %
\usepackage{mathtools} %
\usepackage{booktabs} %
\usepackage{tikz} %
\usepackage{xfrac}

\usepackage{amsmath,amsfonts,bm,amssymb,amsthm}

\definecolor{hanblue}{HTML}{3F88C5}%
\hypersetup{
hidelinks,
    colorlinks=true,
    linkcolor=hanblue,
    urlcolor=hanblue,
    citecolor=hanblue,
    anchorcolor=black}
    
\newtheorem{theorem}{Theorem}[section]
\newtheorem{corollary}[theorem]{Corollary}
\newtheorem{lemma}[theorem]{Lemma}
\newtheorem{proposition}[theorem]{Proposition}
\newtheorem{assumption}[theorem]{Assumption}
\newtheorem{definition}[theorem]{Definition}
\newtheorem{example}[theorem]{Example}

\makeatletter
\newcommand{\printfnsymbol}[1]{%
  \textsuperscript{\@fnsymbol{#1}}%
}
\makeatother

\def\eqref#1{equation~\ref{#1}}

\def\1{\bm{1}}

\def\vf{{\bm{f}}}
\def\vg{{\bm{g}}}
\def\vh{{\bm{h}}}

\def\vx{{\bm{x}}}

\def\mI{{\bm{I}}}

\def\mK{{\bm{K}}}

\DeclareMathAlphabet{\mathsfit}{\encodingdefault}{\sfdefault}{m}{sl}
\SetMathAlphabet{\mathsfit}{bold}{\encodingdefault}{\sfdefault}{bx}{n}

\def\sP{{\mathbb{P}}}

\def\sR{{\mathbb{R}}}

\newcommand{\E}{\mathbb{E}}

\newcommand{\R}{\mathbb{R}}

\DeclareMathOperator*{\argmin}{arg\,min}

\newcommand{\norm}[1]{{\lVert#1\rVert}_2}
\newcommand{\normAny}[1]{{\lVert#1\rVert}}
\newcommand{\ba}{\boldsymbol{a}}

\newcommand{\bs}{\boldsymbol{s}}
\newcommand{\bx}{\boldsymbol{x}}
\newcommand{\by}{\boldsymbol{y}}
\newcommand{\beps}{\boldsymbol{\epsilon}}
\newcommand{\boldeta}{\boldsymbol{\eta}}
\newcommand{\bmu}{\boldsymbol{\mu}}
\newcommand{\bnu}{\boldsymbol{\nu}}
\newcommand{\bsigma}{\boldsymbol{\sigma}}

\newcommand{\bome}{\boldsymbol{\omega}}
\newcommand{\btheta}{\boldsymbol{\theta}}

\newcommand{\bh}{\boldsymbol{h}}

\newcommand{\expvalue}[2]{\underset{#1}{\mathbb{E}}\left[#2\right]}

\newcommand{\calA}{\mathcal{A}}

\newcommand{\calD}{\mathcal{D}}

\newcommand{\calH}{\mathcal{H}}

\newcommand{\calN}{\mathcal{N}}
\newcommand{\calO}{\mathcal{O}}

\newcommand{\calS}{\mathcal{S}}

\newcommand{\calW}{\mathcal{W}}

\newcommand{\calX}{\mathcal{X}}

\newcommand{\supp}[1]{\text{supp}(#1)}

\newcommand{\vspacefigure}{\vspace{-0pt}}
\newcommand{\vspaceparagraph}{\vspace{-0pt}}
\newcommand{\vspaceequation}{\vspace{-0pt}}

\newcommand{\vspacecaption}{\vspace{-0pt}}
\newcommand{\vspacecaptionlow}{\vspace{-0pt}}
\newcommand{\vspacesubcaption}{\vspace{-0pt}}
\newcommand{\vspacesubcaptionlow}{\vspace{-0pt}}
\usepackage{algorithm}
\usepackage{algpseudocode}
\usepackage{caption}
\usepackage{subcaption}
\DeclareMathOperator{\diag}{diag}

\title{Hallucinated Adversarial Control for Conservative Offline Policy Evaluation}

\author[1]{Jonas~Rothfuss\thanks{ Equal contribution.}}
\author[1]{Bhavya~Sukhija\printfnsymbol{1}}
\author[1]{Tobias~Birchler\printfnsymbol{1}}
\author[1]{Parnian~Kassraie}
\author[1]{Andreas~Krause}
\affil[1]{%
    ETH Zurich\\
    Switzerland
}
  
\begin{document}
\maketitle

\begin{abstract}
\looseness -1  We study the problem of {\em conservative off-policy evaluation (COPE)} where given an offline dataset of environment interactions, collected by other agents, we seek to obtain a (tight) lower bound on a policy's performance. This is crucial when deciding whether a given policy satisfies certain minimal performance/safety criteria before it can be deployed in the real world.
To this end, we introduce \textsc{HAMBO}, which builds on an uncertainty-aware learned model of the transition dynamics. To form a conservative estimate of the policy's performance, \textsc{HAMBO} hallucinates worst-case trajectories that the policy may take, within the margin of the models' epistemic confidence regions.
We prove that the resulting COPE estimates are valid lower bounds, and, under regularity conditions, show their convergence to the true expected return. Finally, we discuss scalable variants of our approach based on Bayesian Neural Networks and empirically demonstrate that they yield reliable and tight lower bounds in various continuous control environments.

\end{abstract}

\vspacecaption
 \section{Introduction}
 \vspacecaptionlow
\looseness -1 Reinforcement learning methods require many interactions with their environment to successfully learn and evaluate policies.
Therefore, they are rarely applied in challenging real-world applications such as medicine \citep{murphy2001marginal}, education \citep{mandel2014offline} or autonomous driving \citep{kiran2021deep}, where a policy can only be deployed in the environment if it exceeds a pre-specified performance threshold or fulfills certain safety criteria. This leaves us with a challenging problem: How do we know whether a policy fulfills the necessary criteria so that it can safely interact with the environment, without testing it on the environment, and in the process, compromising safety?

\looseness -1 Off-policy evaluation (OPE) aims to solve this problem by estimating the performance of an evaluation policy, using only offline data that was previously collected by other agents \citep[e.g.][]{precup2001off, dudik2011doubly}.
In practice, offline datasets are often recorded interactions of a human expert with the environment.
Since the evaluation policy typically induces a different action-state distribution than offline data, OPE methods often have to make predictions under strong distribution shifts. As a result, most existing OPE estimators suffer from high variance and are prone to overestimating the performance of the policy \citep{thomas2015hcope}. In safety-critical applications, we can not risk and deploy a policy that is potentially much worse than what the OPE estimate suggests.
Therefore, we aim for {\em conservative off-policy evaluation (COPE)} which seeks a (tight) lower bound on the evaluation policy's expected return that holds with high probability.
Once deployed, the policy may end up exploring areas that were not included in the offline data. Thus, reliably bounding the worst-case performance can be quite challenging.

\looseness -1 We develop a novel {\em model-based} COPE approach that hinges upon two key ideas: {\em epistemic uncertainty} and {\em pessimism}. In particular, our approach, {\em Hallucinated Adversarial Model-Based Off-policy evaluation (\textsc{HAMBO})} (\textsc{HAMBO}), builds on a learned statistical model of the transition dynamics that is able to quantify epistemic uncertainty. To obtain a valid lower bound on the policy performance, HAMBO hallucinates adversarial/worst-case trajectories the agent may take within the epistemic confidence sets of the model.

\looseness-1 We prove that \textsc{HAMBO} reliably yields a high-probability bound on the true expected return of the policy, even when the offline data does not cover the areas explored by the evaluation policy (Proposition~\ref{prop:valid_lower_bound}).
Under regularity conditions, we further show that our conservative estimate {\em converges} from below to the true expected return (Theorem~\ref{theorem:consistency}).
To the best of our knowledge, \textsc{HAMBO} is the first provably consistent and conservative approach for OPE in continuous action-state spaces.
We then propose scalable Bayesian neural network (BNN) variants of \textsc{HAMBO} and empirically evaluate them on various continuous control tasks. Importantly, we demonstrate that, {\em even when the regularity conditions are not met}, \textsc{HAMBO} reliably provides tight lower bounds on the true expected return. \looseness -1
\vspacecaption
\section{Problem Setting}
\vspacecaptionlow
We consider a finite horizon Markov decision process (MDP) $\mathcal{M} = (\mathcal{S}, \mathcal{A}, p_0, p, r, T)$ 
with continuous state and action spaces $\mathcal{S} \subseteq \R^{d_s}$ and $\mathcal{A} \subseteq \R^{d_a}$, initial state distribution $p_0(\bs_0)$, reward function $r(\ba_t, \bs_t)$ and horizon $T \in \mathbb{N}$.
In particular, we consider stochastic transition dynamics that are governed by 
$
    \bs_{t+1} = \vf(\bs_t, \ba_t) + \beps_t
$
where $\vf: \calS \times \calA \rightarrow \calS$ is unknown and $\beps_t \in \R^{d_s}$ is independent, additive transition noise with distribution $p_{\beps}(\beps_t|\bs_t, \ba_t)$. Hence, the transition distribution $p$ follows as
$p(\bs_{t+1}| \bs_t, \ba_t) = p_{\beps}(\bs_{t+1} - f(\bs_t, \ba_t) | \bs_t, \ba_t)$. For simplicity, we assume that the reward function is known. However, all results can straightforwardly be extended to unknown rewards. \looseness-1

The agent interacts with the environment according to a policy $\pi(\ba_t | \bs_t)$, which is a distribution over actions, conditioned on the current state $\bs_t$.
The performance of a policy is typically measured by its expected return $J(\pi) := J_p(\pi) := \mathbb{E}_{\bs_0 \sim p_0}[V_{p, 0}^\pi(\bs_0)]$ where 
$V_t^\pi(\bs) := V_{p, t}^\pi(\bs) := \mathbb{E}_{p,\pi}[G_t|S_t = \bs]$ is the value function and $G_t := \sum_{t'=t+1}^T r(\bs_{t'}, \ba_{t'})$ is the return. For simplicity, we omit a discount factor in the return computation. However, all results presented can be straightforwardly adapted to discounted rewards.
Furthermore, we denote the {\em occupancy measure} of policy $\pi$ as
\begin{equation*}
\vspaceequation
    \rho^{\pi}(\bs, \ba) := \frac{1}{T} \sum_{t=0}^{T-1} p(\bs_t=s, \ba_t=a | \pi, \mathcal{M}) \;,
    \label{eq:state_occ_measure}
\vspaceequation
\end{equation*}
that is, the probability density function of being in state $\bs$ and performing action $\ba$ at any point of time $t = 0, ..., T-1$.

We study the problem of offline policy evaluation where the task is to evaluate the performance, i.e. estimate the expected return $J(\pi_e)$, of a given evaluation policy $\pi_e$ while only using an offline dataset $\calD_b = \{(\bs_i, \ba_i, r_i, \bs'_i)\}_{i=1}^n$
of observed transitions.
The key challenge in OPE is the distribution shift between the (unknown) behavior policy $\pi_b$ which generated the dataset $\calD_b$ and the policy $\pi_e$ which we would like to evaluate. If $\pi_b$ differs from $\pi_e$, their state occupancy measures $\rho^{\pi_b}$ and $\rho^{\pi_e}$ can look significantly different. As a result, the dataset $\calD_b$ which is generated based on $\rho^{\pi_b}$ may contain many samples in regions of the state-action space which $\pi_e$ is unlikely to visit and limited data in regions that are relevant for accurately evaluating $\pi_e$. In some cases, the support of $\rho^{\pi_b}$ might not even contain the support of $\rho^{\pi_e}$, i.e., 
$\exists (\bs, \ba) \in \mathcal{S} \times \mathcal{A}: \rho^{\pi_e}(\bs, \ba) > 0 \wedge \rho^{\pi_b}(\bs, \ba) = 0$. 
Since OPE methods have to make predictions under such strong distribution shifts their estimates suffer from high variance and are prone to overestimate the performance of the policy.

OPE is particularly relevant in applications where we need to ensure a certain level of performance before a policy can be deployed online. Hence, it is often important to reliably determine whether or not the policy $\pi_e$ meets its minimum performance requirements. We formalize this problem as {\em conservative offline policy evaluation} (see Definition \ref{definition:conservativeope}) where we want to ideally find a tight lower bound on the expected return that holds with high-probability:\looseness-1

\begin{definition}[Conservative Offline Policy Evaluation]\hfill\\
\label{definition:conservativeope}
Let $\mathcal{M}$ be an MDP and $\calD_b \in (\mathcal{S} \times \mathcal{A} \times \mathbb{R} \times \mathcal{S})^n$ a dataset of transitions, collected with a behavior policy $\pi_b$ on $\mathcal{M}$.
Then the task of conservative OPE is:
Given the offline dataset $\calD_b$, a policy $\pi_e$ to evaluate and a confidence level $\delta \in (0,1)$, find the largest possible lower-bound $b \in \mathbb{R}$, which satisfies $b \leq J(\pi_e)$ with probability at least $1-\delta$.
\vspaceparagraph
\end{definition}
\vspaceparagraph
\looseness -1 In some applications \citep[e.g.,][]{brunke2022safe}, safety criteria are not directly encoded in the reward and instead, are expressed as additional constraints in the form of $\E_{(\bs, \ba)\sim \rho^{\pi_e}} \left[ c_i(\bs, \ba)\right] \geq 0$. To determine with high confidence whether $\pi_e$ meets these constraints, we can apply COPE to each $c_i$ individually.

\vspacecaption
\section{COPE via Adversarial Transition Models}
\vspacecaptionlow
\label{section:hambo_general_idea}
We take a model-based approach to COPE, and use a statistical model to estimate which transition functions $\bh: \calS \times \calA \rightarrow \calS$ from a hypothesis space $\calH$ are plausible given the offline data $\calD_b$ of size $n$. Then, we employ this statistical model of the transition dynamics to estimate the policy value $J(\pi_e)$.
For this estimate, we want to guarantee with high probability that it does not exceed the true policy value. To ensure this, we need to be able to reliably quantify the {\em epistemic uncertainty} of our model estimates.

\looseness -1 Uncertainty quantification can be done with either a frequentist approach that produces mean and confidence estimate $\bmu_n(\bs, \ba)$ and $\bsigma_n(\bs, \ba)$ or with a Bayesian model that maintains a posterior distribution $p(\vh|\calD_b)$ over dynamics models in $\calH$. In the Bayesian case, we denote $\bmu_n(\bs, \ba):= \E_{\vh \sim p(\bh|\calD_b)} [\bh(\bs, \ba)]$ as the posterior mean and $\bsigma^2_n(\bs, \ba) :=\diag(\E_{\vh, \vh' \sim p(h|\calD_b)}[\vh(\bs, \ba) \vh'(\bs, \ba)^\top])$ as the posterior variance. In either case, we require that our statistical model of $\bh$ is calibrated:

\begin{assumption}[Calibrated model] \label{assumption:calibration}
    A statistical model $(\bmu_n, \bsigma_n, \beta_n)$, with $\beta_n(\delta) \in \R^+$ as a scalar function that depends on the confidence level $\delta \in (0, 1]$, is calibrated with respect to $\vf$ if, with probability at least $1-\delta$, for all $ (\bs, \ba) \in \calS \times \calA$ and $j=1, \dots, d_s$
    \vspaceequation
    \begin{equation*}
    \vspaceequation
        |\mu_{n,j}(\bs, \ba) - f_{j}(\bs, \ba) | \leq \beta_n(\delta) \sigma_{n,j}(\bs, \ba),
        \vspaceequation
    \end{equation*}
    where $\mu_{n,j}$ and $\sigma_{n,j}$ denote the $j$-th element in the vector-valued functions $\bmu_n$ and $\bsigma_n$, respectively.
\end{assumption}
\vspaceparagraph
\looseness -1 Popular statistical models for transition dynamics that capture epistemic uncertainty are {\em Gaussian Processes (GPs)} \citep{rasmussen2005gp},  {\em Probabilistic Neural Network Ensembles} \citep{lakshminarayanan2017uncertaintyensembles} and {\em Bayesian Neural Networks} \citep{blundell2015bnn}. In later sections, we will attend to these specific choices of model in more detail and discuss when they are calibrated.\looseness-1
\vspacecaption
\subsection{The \textsc{HAMBO} Framework}
\vspacecaptionlow
\vspace{-1mm}
 \label{subsec:general_hambo_approach}
\looseness-1 If our model is calibrated, we can, with high probability, use the confidence region 
\vspaceequation
\[
\vspaceequation
[\bmu_n(\bs, \ba) - \beta_n(\delta) \bsigma_n(\bs, \ba), \bmu_n(\bs, \ba) + \beta_n(\delta) \bsigma_n(\bs, \ba)]
\vspaceequation
\]
which is a $d_s$-dimensional hypercube, as a proxy for the true dynamics $\vf(\bs, \ba)$. We then pessimistically select transitions within this region, to guarantee a high probability lower bound on the policy value $J(\pi_e)$. We do so, by introducing an adversary $\boldeta: \calS \times \calA \rightarrow [-1,1]^{d_s}$ that, for every $(\bs, \ba) \in \calS \times \calA$ picks a transition from the confidence region, thereby inducing the following hallucinated transition distribution:
\begin{align} \label{eq:hallucinated_model}
\begin{split}
    \tilde{p}_{\boldeta}(\bs_{t+1}| \bs_t, \ba_t) := p_{\beps}\big(&\bs_{t+1} - \bmu_n(\bs_t, \ba_t)\\
    &- \beta_n \boldeta(\bs_t, \ba_t)  \bsigma_n(\bs_t, \ba_t) \big). 
\end{split}
\end{align}
This allows us to obtain a conservative value estimate for $\pi_e$\looseness-1
\begin{equation} \label{eq:pessimistic_value}
\vspaceequation
    \tilde{J}(\pi_e) := \min_{\boldeta} J_{\tilde{p}_{\boldeta}}(\pi_e) ~.
    \vspaceequation
\end{equation}
This equation summarizes our approach {\em hallucinated adversarial model-based off-policy evaluation (\textsc{HAMBO})} and Algorithm~\ref{alg:hambo_basic} presents the pseudo-code. 
Here, the expected reward $J_{\tilde{p}_\eta}(\pi_e)$ of $\pi_e$ under the hallucinated transition model $\tilde{p}_\eta$ can, e.g., be estimated via Monte Carlo estimation (i.e., generating trajectory rollouts and averaging the respecting returns). To find the adversary $\boldeta(\bs, \ba)$ which minimizes (\ref{eq:pessimistic_value}), we can view $\boldeta(\bs, \ba)$ as policy that aims to maximize $- J_{\tilde{p}_\eta}(\pi_e)$ and solve the corresponding optimal control problem. Importantly, with high probability, $\tilde{J}(\pi_e)$ is a lower bound on the true policy value $J(\pi_e)$:

\begin{proposition}[Valid lower bound] \label{prop:valid_lower_bound} 
Given a calibrated model $(\bmu_n, \bsigma_n, \beta_n(\delta))$, the \textsc{HAMBO} estimates satisfy $\tilde{J}(\pi_e) \leq J(\pi_e)$, with probability greater than $1-\delta$.
\end{proposition}
\vspaceparagraph
While Proposition \ref{prop:valid_lower_bound} shows that our estimate $\tilde{J}(\pi_e)$ fulfills the requirements of COPE, $\tilde{J}(\pi_e)$ could potentially be very loose. However, 
 we can further establish a worst-case lower bound on $\tilde{J}(\pi_e)$, if $\vf$, $r$, $\bsigma_n$ and $\pi_e$ are continuous. Formally, we make the following Lipschitz continuity assumption:

\begin{assumption} (Lipschitz continuity) \label{assumption:lip_continuity}
  $\vf$ is $L_f$-Lipschitz, $r$ is $L_r$-Lipschitz, $\bsigma$ is $L_\sigma$-Lipschitz and $\pi_e$ is $L_\pi$-Lipschitz w.r.t. the Wasserstein-1 distance, i.e., for all $\bs, \bs' \in \calS$
  \vspaceequation
  \begin{equation}
  \vspaceequation
      \calW_1\hspace{-2pt}\left(\pi(\ba|\bs), \pi(\ba|\bs') \right) \leq L_\pi \norm{\bs - \bs'}.
      \vspaceequation
  \end{equation}
\end{assumption}
Here, the continuity assumption on $\pi$ is expressed in terms the Wasserstein-1 distance and implies that a small change in the state space only induces a proportionally small change in the conditional action distribution of the policy. For instance, this is the case for policies that can be reparametrized with a Lipschitz function which is very common in practice:
\begin{example} \label{example:reparametrizable_policy_is_w1_lip}
\looseness -1 Any policy $\pi(\ba|\bs)$ that can be reparametrized as $\vg(\bs, \boldsymbol{\zeta})$, where $\boldsymbol{\zeta} \sim p(\boldsymbol{\zeta})$ and $\vg$ is $L_g$-Lipschitz, is also $L_g$-Lipschitz w.r.t. the $\calW_1$- distance.
\vspaceparagraph
\end{example}
\vspaceparagraph
Such Lipschitz assumptions are common in model-based OPE \citep[e.g.][]{fontenau2009lower_bound, paduraru2013off} and RL more broadly \citep[e.g.][]{berkenkamp2017safe, curi2020hucrl}, and, e.g, hold in many real-world control problems.
With these regularity assumptions, we bound how far away the \textsc{HAMBO} estimate $\tilde{J}(\pi_e)$ is from the true policy value: 

\begin{theorem} \label{theorem:lower_bound}
Under Assumption \ref{assumption:calibration} and \ref{assumption:lip_continuity} we have, with probability at least $1-\delta$, that
\vspaceequation
    \begin{equation*}
    \vspaceequation
       J(\pi_e) - \tilde{J}(\pi_e) \leq  C_n ~ \expvalue{(\bs, \ba) \sim \rho^{\pi_e}}{ \norm{\bsigma_n(\bs, \ba)}}
       \vspaceequation
    \end{equation*}
    \vspaceequation
    where 
    \vspaceequation
    \[
    \vspaceequation
    C_n \hspace{-2pt} \coloneqq \hspace{-2pt} \Bar{L}_r \left(1\hspace{-2pt} +\hspace{-2pt} \sqrt{d_s}\right)\beta_n T^2 
 \hspace{-2pt} \left(1\hspace{-2pt} +\hspace{-2pt} \Bar{L}_f \hspace{-2pt}  + \hspace{-2pt} (1\hspace{-2pt} +\hspace{-2pt} \sqrt{d_s})\beta_n(\delta)\Bar{L}_{\sigma}\right)^{\hspace{-2pt} T-1}  \hspace{-10pt}
 \]
    with $\Bar{L}_r := L_r (1+L_\pi)$ and $\Bar{L}_f, \Bar{L}_\sigma$ defined analogously.
    \vspaceparagraph
\end{theorem}
\vspaceparagraph
This theorem shows how by tuning the confidence level $\delta$, we can trade-off accuracy with reliability. In particular, choosing a small $\delta$ will ensure that the upper-bound on $J(\pi_e)$ holds. However, it also increases $\beta_n(\delta)$ and loosens the bound, indicating the $\tilde J(\pi_e)$ estimate will be less accurate.
Tightness of the HAMBO lower bound $\tilde{J}(\pi_e)$ depends on the
following key factors: Lipschitz-regularity, episode horizon $T$, and epistemic uncertainty.
Mainly, smaller Lipschitz constants and shorter episode lengths improve the bound. Moreover, the smaller the expected epistemic standard deviation $\bsigma_n(\bs_t, \ba_t)$ under the state occupancy measure of $\pi_e$, the tighter the bound. 
While the first two factors are generally dictated by the problem instance, the epistemic uncertainty can be reduced by using more offline data (in the relevant areas of the state-action space). If we can show that the epistemic uncertainty shrinks sufficiently fast with the number of offline data points $n$ (i.e., faster than $\calO(\beta_n^{T})$), then we can prove that $\tilde{J}(\pi_e)$ converges to the true policy value as $n \rightarrow \infty$. 
In the following, we discuss corresponding sufficient convergence conditions for GP models.

\subsection{\textsc{HAMBO} with Smooth GP functions}
 \vspacecaptionlow
 \label{section:hambo_gp_models}
In this section, we discuss the application of GPs for constructing calibrated confidence regions to be used for \textsc{HAMBO}. %
For the transition dynamics, we consider vector-valued functions $\vf(\bs, \ba) \mapsto (f_1(\bs, \ba), ..., f_{d_s}(\bs, \ba))$ such that the scalar-valued functions $f_j \in \calH_k$ reside in a Reproducing Kernel Hilbert Space (RKHS) $\calH_k$ with kernel function $k(\cdot, \cdot)$ and have bounded RKHS norm, i.e. $\normAny{f_j}_k \leq B$. 
We denote this space by  \smash{$\vf \in \calH_{k, B}^{d_s} = \{ [f_1, ..., f_{d_s}] : \normAny{f_j}_k \leq B, j=1, ..., d_s \}$}.
We assume that transition noise $\beps \sim \calN(0, \sigma^2_\epsilon \bm{I})$ is normally distributed with variance $\sigma^2_\epsilon$.

By fitting a zero-mean Gaussian Process $\mathcal{GP}(0, k)$ on each dimension $j=1, ..., d_s$ of the next state $\bs_{t+1}$, we can use the posterior means and variances to construct calibrated confidence sets. For brevity, we denote $\bx := (\bs, \ba)$, so that
\vspaceequation
\begin{align}
\begin{split}
    \mu_{n,j} (\bx)& = {\bm{k}}_{n}^\top(\bx)({\bm K}_{n} + \sigma^2_\epsilon \bm{I})^{-1}\by_{n, j}  \label{eq:GPposteriors}\\
     \sigma^2_{n, j}(\bx) & =  k(\bx, \bx) - {\bm k}^T_{n}(\bx)({\bm K}_{n}+\sigma^2_\epsilon \bm{I})^{-1}{\bm k}_{n}(\bx)
\end{split}
\end{align}
where $\by_{n, j} = [s'_{i, j}]^\top_{i \leq n}$ is the vector the $j$-th element of the observed next states $\bs'_i$, $\bm{k}_{n}(\bx) = [ k(\bx, \bx_i)]^\top_{i \leq n}$, and ${\bm K}_{n} = [ k(\bx_i, \bx_l)]_{i,l \leq n}$ is the kernel matrix. By concatenating the element-wise posterior mean and standard deviation, we obtain $\bmu_n(\bx) = [\mu_{n,j}(\bx)]^\top_{j \leq d_s}$ and
 $\bsigma_n(\bx) = [\sigma_{n,j}(\bx)]^\top_{j \leq d_s}$. Using this, we can construct calibrated confidence intervals that fulfill Assumption \ref{assumption:calibration}:

\begin{lemma}[Calibrated GP confidence sets] \label{lem:gp_calibrated}
Let $\vf \in \calH_{k,B}^{d_s}$.
Suppose $\bmu_n$ and $\bsigma_n$ are the posterior mean and variance of a GP with kernel $k$, fitted to $n$ noisy evaluations of $\vf$.
There exists $\beta_n(\delta)$, for which the tuple $(\bmu_n, \bsigma_n, \beta_n(\delta))$ satisfies Assumption~\ref{assumption:calibration} w.r.t. function $\vf$.
\end{lemma}
\vspaceparagraph

\looseness -1 In Appendix~\ref{app:gp_proofs} we prove this lemma using results of \citet{chowdhury2017kernelized} and give the exact expression for a $\beta_n(\delta)$ that satisfies it. Generally, $\beta_n(\delta)$ depends on the {\em maximum information capacity} $\gamma_n$ of the kernel (see Appendix \ref{app:gp_proofs} for definition and details).
In the GP setting, we can also show Lipschitz continuity of $\vf$ and $\bsigma$, if the kernel function $k$ is sufficiently regular:
\begin{lemma} \label{lemma:lip_f_sigma_gp}
    If the kernel metric $d_k(\bx, \bx') := (k(\bx, \bx) + k(\bx', \bx') - 2 k(\bx, \bx'))^{\frac{1}{2}}$ is $L_k$-Lipschitz, then every $\vf \in \calH_{k, B}^{d_s}$ is Lipschitz with $L_f = \sqrt{d_s} B L_k$ and the posterior standard deviation $\bsigma$ is Lipschitz with $L_\sigma = \sqrt{d_s} L_k$.
\end{lemma}
\vspaceparagraph
For common kernels, the kernel metric is Lipschitz continuous, and thus Lemma \ref{lemma:lip_f_sigma_gp} applies. For instance, for the linear kernel we have $L_k = 1$, for the RBF kernel we have $L_k = 1 / \ell$ and for the Matern-$\nu$ kernel we have $L_k = \sqrt{\nu / (\nu -1)} / \ell$, where $\ell$ is the lengthscale and $\nu$ the smoothness parameter of the Matern kernel.

\looseness -1 We can conclude that conditions of Proposition~\ref{prop:valid_lower_bound} and Theorem~\ref{theorem:lower_bound} are met when a GP is used for learning the transition dynamics from offline data. Hence, when the reward and the policy are Lipschitz, the \textsc{HAMBO} estimate satisfies
\[
\vspaceequation
J(\pi_e) - C_n \expvalue{\rho^{\pi_e}}{ \norm{\bsigma_n(\bs, \ba)}} \leq \tilde J(\pi_e) \leq J(\pi_e)
\vspaceequation
\]
with high probability. We can show that given a dataset of i.i.d.~trajectories, the difference term shrinks with $n$ sufficiently fast:\looseness-1

\begin{theorem}[Consistency of  \textsc{HAMBO}]  \label{theorem:consistency}
    Let $r$ be $L_r$-Lipschitz, $\pi$ be $L_\pi$-Lipschitz w.r.t. the $\calW_1$-distance and $\vf \in \calH_{k, B}^{d_s}$ where $k$ is a kernel with a $L_k$-Lipschitz kernel metric with a maximum information capacity $\gamma_n$ which is $\calO(\mathrm{polylog}(n))$.%
     Suppose both $\rho^{\pi_e}$ and $\rho^{\pi_b}$ have a compact support and 
    $\supp{\rho^{\pi_e}} \subseteq \supp{\rho^{\pi_b}}$ and $\calD_{b}$ consists of $n$ data points from i.i.d. trajectories according to the behavior policy $\pi_b$. Then as $n \rightarrow \infty$,
    \[
    \tilde{J}_n(\pi_e) \xrightarrow[]{\mathrm{a.s.}} J(\pi_e).
    \]   
\end{theorem}
  \vspaceparagraph
\looseness -1 The theorem implies that $\tilde J(\pi_e)$ is not only a conservative estimator for $J(\pi_e)$, but under certain regularity conditions, it is also a consistent estimator of the policy's true value. Meaning that for large $n$, the trade-off between reliability and accuracy vanishes.
In Appendix~\ref{app:consistency_proof} we prove this theorem and give the exact rate at which the \textsc{HAMBO} estimate converges to the true value of $\pi_e$. This rate depends on the choice of kernel, time horizon $T$, and dimensions of the environment $(d_a, d_s)$. As an example, if $k$ is a Linear or RBF kernel, with high probability
$\vert \tilde{J}_n(\pi_e) - J(\pi_e)\vert = \tilde \calO\left(n^{-1/2}\right)$
where $\tilde O$ omits polylogarithmic factors.

\looseness -1 The kernel assumptions in Theorem \ref{theorem:consistency} hold for many popular kernels such as any inner product kernel $k(\bx, \bx') = \phi(\bx)^\top\phi(\bx')$ with Lipschitz continuous finite-dimensional feature maps $\phi(\cdot)$ or smooth kernels such as the RBF. On the other hand, Theorem \ref{theorem:consistency} does hold for non-smooth functions, e.g., those corresponding to a Matern since their maximum information capacity $\gamma_n$ grows polynomially with $n$.

To the best of our knowledge, Theorem~\ref{theorem:consistency} is the first result that shows the consistency of a model-based finite-horizon OPE method for a continuous environments.

\vspacecaption

\section{HAMBO with Neural Networks}
\vspacecaptionlow

\label{sec:hambo_nn}
In practice, we often want to evaluate policies in settings where the state and action spaces are higher-dimensional, and have access to larger amounts of offline data. In such environments, GPs become unpractical as they tend to generalize poorly in high-dimensional domains and their inference becomes prohibitively expensive for larger datasets.

\textbf{The NN-based Statistical Model.} In this section, we discuss practical variants of \textsc{HAMBO} which employ neural networks that scale more favorably to large datasets and high-dimensional domains. Crucially, we need to be able to quantify epistemic uncertainty. For this purpose, we employ Bayesian Neural Networks (BNNs) which model $\bh_{\btheta}(\bs, \ba)$ as a neural network function where $\btheta$ are the parameters of the neural network. BNNs presume a prior distribution $p(\btheta) = \calN(\btheta ; 0, \lambda \bm{I})$ and maintain an approximation of the posterior $p(\btheta| \calD_b) \propto p(\calD_b | \btheta) p(\btheta)$ over neural network parameters. We use an independent Gaussian likelihood $p(\calD_b | \btheta) = \prod_{i=1}^n \calN(\bs_i'; \bh_{\btheta}(\bs_i, \ba_i), \bnu^2_{\btheta}(\bs_i, \ba_i))$ where 
$\bnu^2_{\btheta}(\bs, \ba)$ is the vector of transition noise variances which is also predicted by the BNN.%

We use Stein Variational Gradient Descent (SVGD) \citep{liu2016svgd} to approximate the posterior as a set of $K$ particles $\Theta = \{ \btheta_1, ..., \btheta_K \}$. 
We form the mean prediction of our model as the average prediction of the $K$ NNs:
\vspaceequation
\begin{align*}
\vspaceequation
    \bmu_\Theta(\bs, \ba) &= \frac{1}{K} \sum_{k=1}^K \bh_{\btheta_k}(\bs, \ba). 
\vspaceequation
\end{align*}
Similarly, we estimate the epistemic variance as
\vspaceequation
\begin{align*}
\vspaceequation
    \bsigma^2_{\Theta, e}(\bs, \ba) &= \frac{1}{K} \sum_{k=1}^K (\bh_{\btheta_k}(\bs, \ba) - \bmu_\Theta(\bs, \ba))^2.
    \vspaceequation
\end{align*}
The overall predictive distribution is the equally weighted mixture of all $K$ NN-based conditional Gaussians, i.e.,  
\begin{equation} \label{eq:predictive_gmm}
\vspaceequation
    p(\bs'|\bs, \ba, \calD_b) = \frac{1}{K} \sum_{k=1}^K \calN(\bs'; \bh_{\btheta_k}(\bs, \ba), \bnu^2_{\btheta_k}(\bs, \ba))
    \vspaceequation
\end{equation}
whose variance is $\bsigma_\Theta^2(\bs, \ba) =  \bsigma^2_{\Theta, e}(\bs, \ba) + \bsigma^2_{\Theta, a}(\bs, \ba)$, where \smash{$\bsigma^2_{\Theta, a}(\bs, \ba) := \frac{1}{K}\sum_{k=1}^K \bnu^2_{\btheta_k}(\bs, \ba)$} represents aleatoric and $\bsigma^2_{\Theta, e}(\bs, \ba)$ the epistemic uncertainty. \looseness-1

\looseness -1 \textbf{Calibrating the Model.} Since our BNN model uses approximate inference and a potentially misspecified prior, it may not satisfy the calibration condition of Assumption \ref{assumption:calibration}. Thus, we re-calibrate the model's uncertainty estimates with a calibration set $\calD_c \subset \calD_b$ that is withheld from the training. In particular, we use temperature scaling which chooses $\tau > 0$ such that the scaled predictive distribution (\ref{eq:predictive_gmm}) with variance $\tau^2 \bsigma_\Theta^2(\bs, \ba)$ has a minimal empirical calibration error on $\calD_c$ \citep{kuleshov2018calibration}. Algorithm~\ref{alg:BNN_calibrate} formalizes this technique. Note that re-calibrating the BNN model does not guarantee formal calibration in the sense of Assumption \ref{assumption:calibration}. However, in our experiments, we found it to reliably yield a conservative value estimate $\tilde{J}(\pi_e)$.

\vspacesubcaption

\subsection{Practical NN-Based HAMBO Variants} %
\vspacesubcaptionlow

\label{sec:variants}
In the following, we discuss three ways of constructing adversarially hallucinated transition models based on our BNN model described in Equation~(\ref{eq:predictive_gmm}). The formal pseudo-code of all algorithms is presented in Appendix~\ref{app:algos}.

\looseness -1 \textbf{Continuous Adversary (\textsc{HAMBO-CA}).} This approach directly reflects the hallucinated adversarial transition model, introduced in (\ref{eq:hallucinated_model}) and (\ref{eq:pessimistic_value}). The adversary $\boldeta(\bs, \ba) \in [-1, 1]^{d_s}$ chooses the mean of the Gaussian transition probability from the epistemic confidence set, i.e.,
\begin{align*}
\resizebox{0.99\hsize}{!}{%
$\tilde{p}_{\boldeta}(\bs'| \bs, \ba) \coloneqq \calN \big(\bs';  \bmu_\Theta(\bs, \ba) +  \tau^2 \boldeta(\bs, \ba)  \bsigma^2_{\Theta, e} , \bsigma^2_{\Theta, a}(\bs, \ba) \big) .$
}
\end{align*}
\looseness -1 To get the corresponding conservative value estimate $\tilde{J}(\pi_e)$, we need to solve the minimization problem $\min_{\boldeta} J_{\tilde{p}_{\boldeta}}(\pi_e)$. For this, we parameterize the adversary  $\boldeta(\bs, \ba)$ as a neural network policy and use Soft Actor-Critic (SAC)  \citep{haarnoja2018sac2} to maximize the negative return. 

\textbf{Discrete Adversary (\textsc{HAMBO-DA}).}
\looseness -1 Our BNN posterior is approximated by a set of $K$ NNs whose mean squared error difference corresponds to epistemic uncertainty. Thus, we can also construct a pessimistic transition model by letting the adversary choose which of the $K$ NNs to pick. In this case, the adversary $\vartheta(k | \bs, \ba)$ is a categorical distribution over the NN indices $\{1, ..., K\}$. The hallucinated transition model follows as: 
\vspaceequation
\begin{align*}
\vspaceequation
\tilde{p}_{\vartheta}(\bs'| \bs, \ba) :=  \sum_{k=1}^K \vartheta(k | \bs, \ba) \calN \big(\bs' ;  \bh_{\btheta_k}(\bs, \ba) , \bnu^2_{\btheta_k}(\bs, \ba) \big).
\vspaceequation
\end{align*}
Here, the adversary stochastically picks one of NN models at every step $t=0, ..., T-1$. For this reason, we refer to this variant as $\textsc{DA1}$ (Algorithm~\ref{alg:hambo_da-1}). The corresponding value estimate follows as $\tilde{J}_{\textsc{DA1}}(\pi_e) = \min_{\vartheta} J_{\tilde{p}_{\vartheta}}(\pi_e)$. We solve the optimization problem by parameterizing the adversary $\vartheta$ as a NN policy and use the clipped double DQN algorithm \cite{fujimoto2018td3} to maximize the negative return.

Alternatively, we can constrain the adversary so that it has to commit to one of the $K$ NN models for the entire trajectory. We refer to this variant as $\textsc{DAinf}$ (Algorithm~\ref{alg:hambo_dainf}). In this case, the transition model corresponds to the predictive distribution of one of the NNs $p_{\btheta_k}(\bs'| \bs, \ba) =  \calN \big(\bs' ;  \bh_{\btheta_k}(\bs, \ba), \bnu^2_{\btheta_k}(\bs, \ba) \big)$, and the value estimate follows as the minimum the policy values under each of the models, i.e.,  $\tilde{J}_{\mathrm{DAinf}}(\pi_e) = \min_{k \in \{1, \dots, K\}} J_{p_{\btheta_k}}(\pi_e)$.
If $K$ is larger (e.g., $K > 20$), we recommend taking the empirical $\delta$ quantile of the policy values $\{J_{\bar{p}_{k}}(\pi_e)\}_{k=1}^K$ instead of the minimum. In this case, $\textsc{DAinf}$ has similarities to the model-based bootstrap approach of \citet{kostrikov2020bootstrap}. \looseness -1

Naturally, the value estimates of $\textsc{DAinf}$ are less pessimistic than those of $\textsc{DA1}$, i.e. $\tilde{J}_{\mathrm{DAinf}}(\pi_e) \geq \tilde{J}_{\mathrm{DA1}}(\pi_e)$, because the adversary cannot change which model it picks throughout the trajectory. In the experiment section, we investigate whether $\textsc{DAinf}$ is still conservative enough to reliably yield lower bounds on the true policy values $J(\pi_e)$.

\begin{figure}[ht]
    \centering
    \includegraphics[width=\columnwidth]{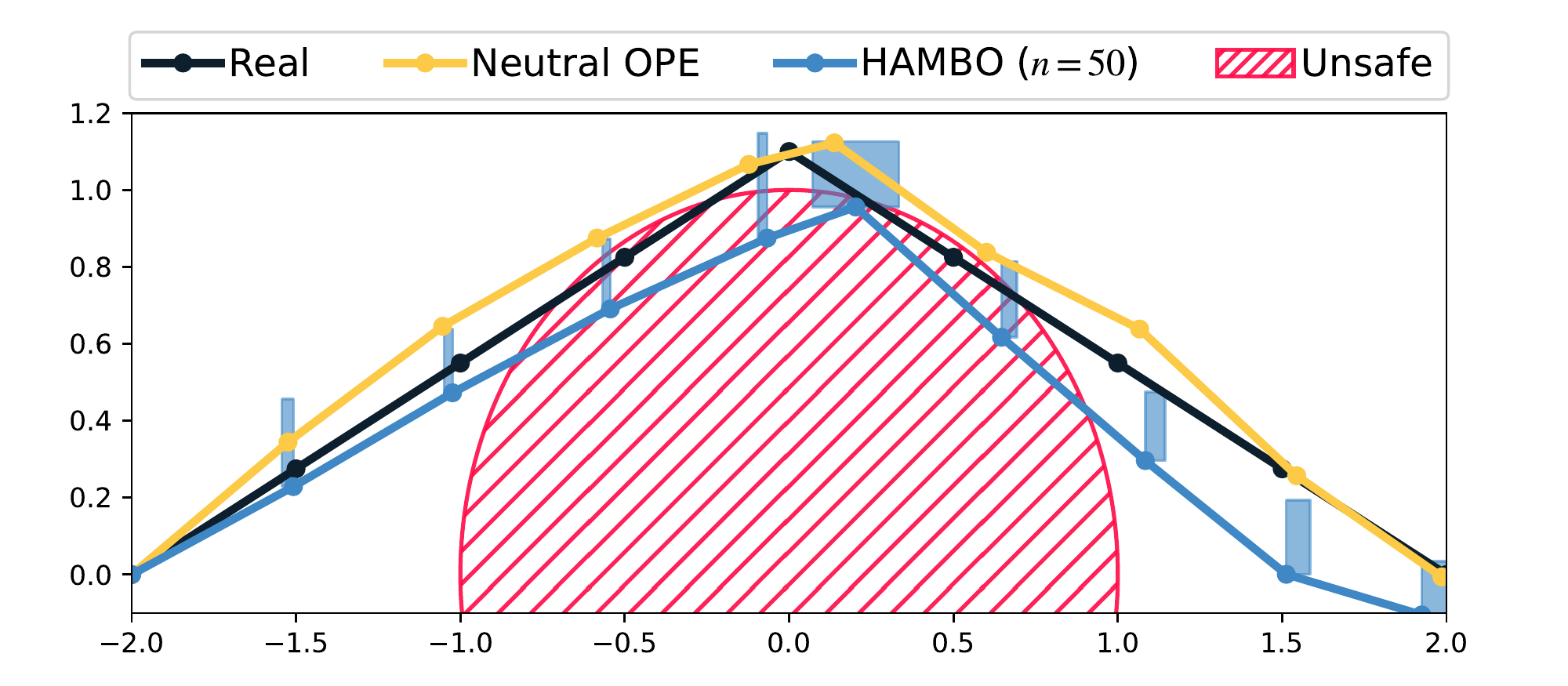}
    \vspacefigure
    \vspacecaption
    \caption{Hallucinated trajectories for model-based OPE and pessimistic \textsc{HAMBO}. While OPE overestimates the performance of the unsafe policy, \textsc{HAMBO} correctly gives a conservative estimates through its adversarial transition model. The adversary chooses the worst-case trajectory with the confidence sets (shaded blue areas). }\label{fig:toyexample}
    \vspacefigure
\end{figure}
\vspacecaption
\vspace{-3mm}
\section{Experiments}
\vspacecaptionlow
\vspace{-1mm}
\label{sec:results}

\looseness-1 We start this section by illustrating the inner workings of \textsc{HAMBO} with a toy example to show why pessimism is crucial for COPE.
We demonstrate that the convergence guarantees from Section~\ref{section:hambo_gp_models} materialize in practice for GP models.
Finally, we empirically evaluate and compare the practical variants of \textsc{HAMBO} with BNNs on various continuous control tasks. For comparability between our environments, we shift and scale all our results so that the true policy return value $J(\pi_e)$ is 1. 

\begin{figure*}[ht]
    \centering
        \includegraphics[width=0.38\textwidth, trim={5mm 3mm 2mm 1mm},clip]{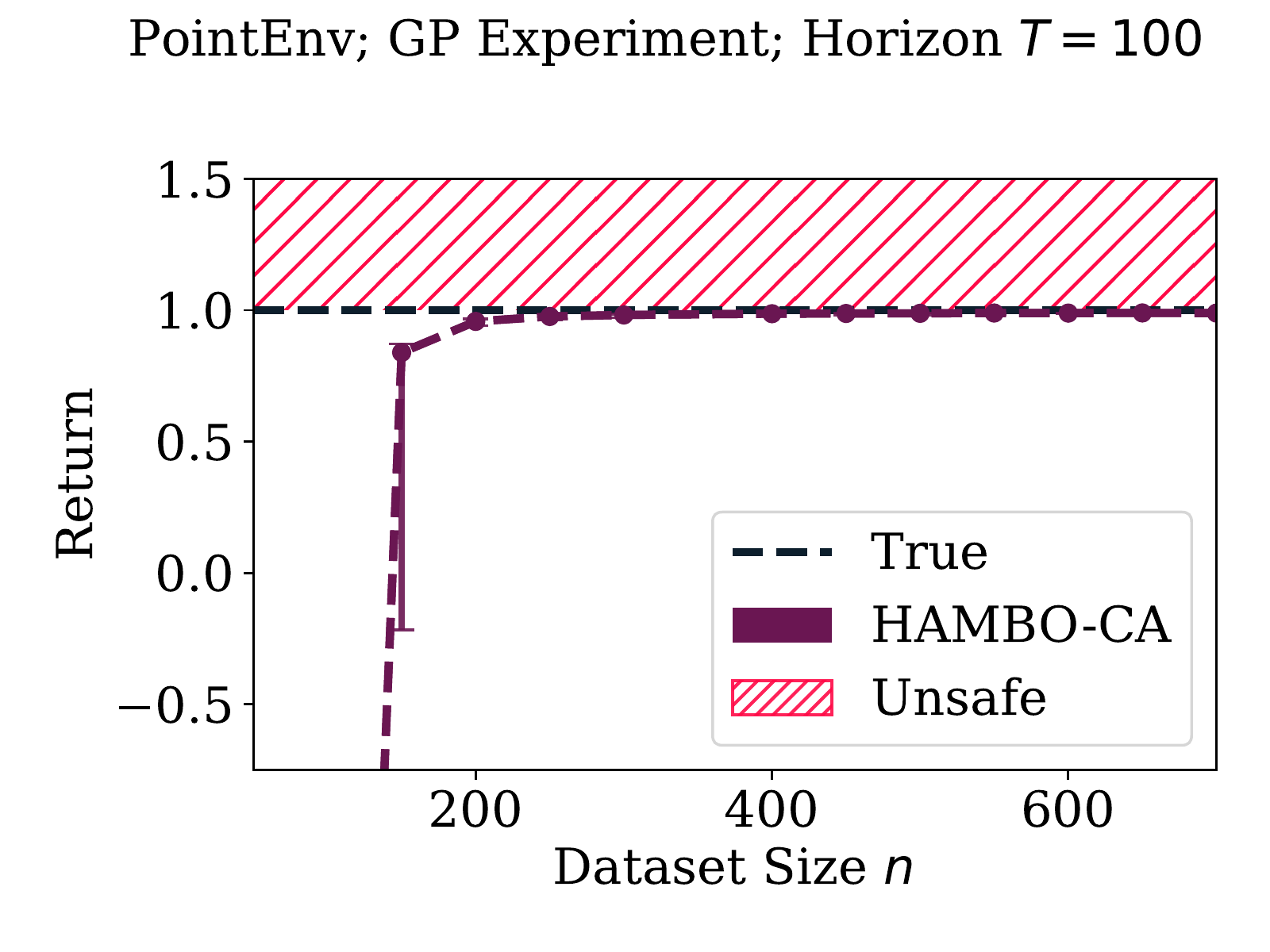}
        \hspace{20pt}
        \includegraphics[width=0.38\textwidth, trim={5mm 3mm 2mm 1mm}, clip]{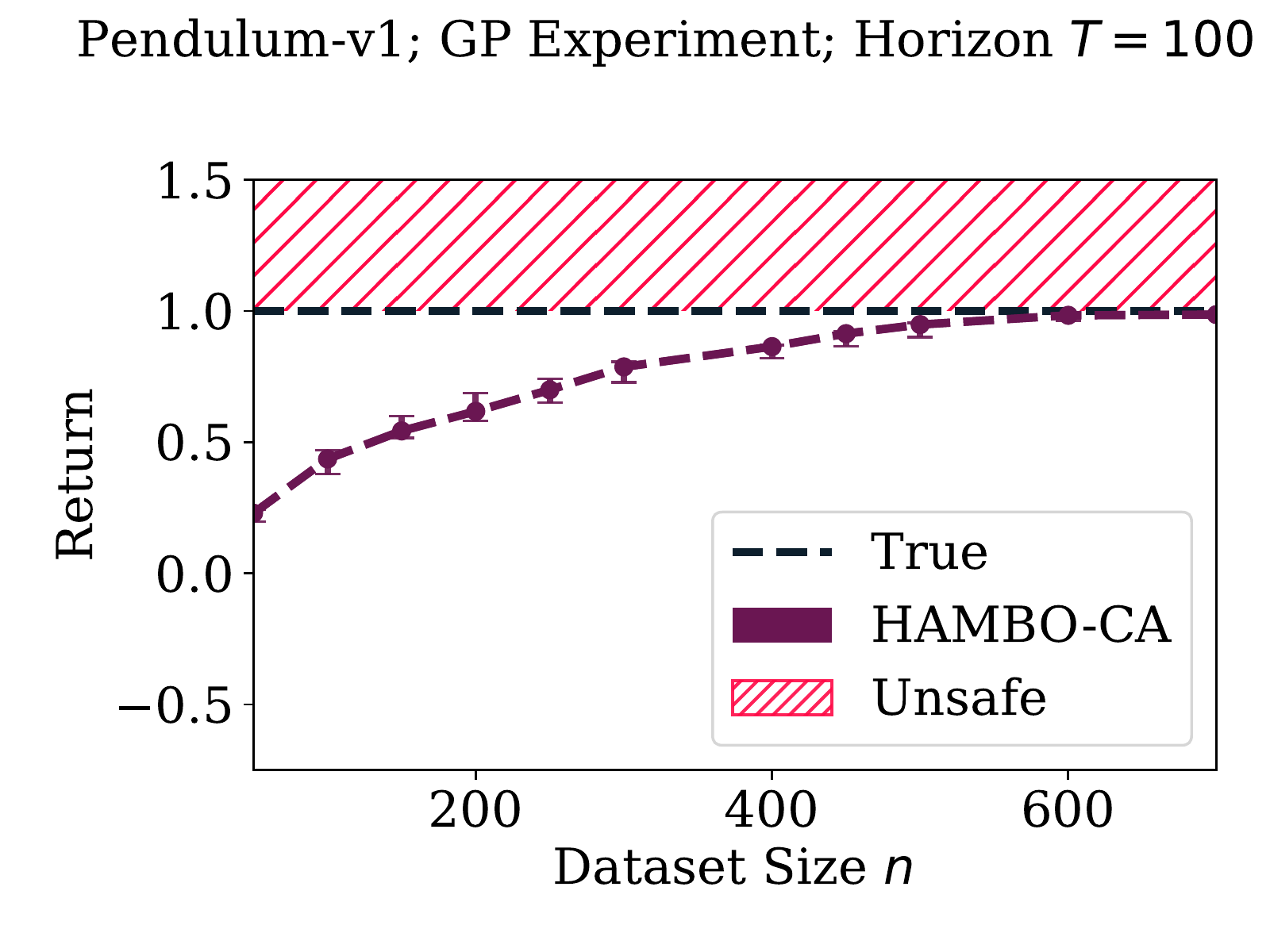}
\vspace{-4pt}
    \caption{\looseness -1 GP-based \textsc{HAMBO} for increasing offline dataset sizes $n$ evaluated on the PointEnv and Pendulum-v1. The lower bound approaches the true return.\label{fig:gp_consistency} \vspace{10pt}}    
\vspacefigure
\vspacefigure
\end{figure*}
\vspace{-3mm}
\subsection{Illustrative Example}
\vspacecaptionlow
\vspace{-1mm}
To illuminate the core idea of \textsc{HAMBO} and  why pessimism is crucial for COPE, we conduct experiments on a toy environment which we call PointSafety (see Figure~\ref{fig:toyexample}).
In this environment, the agent navigates in the two-dimensional plane by applying actions $\ba \in [-0.5, 0.5]^2$ such that its position (i.e, state $\bs \in \calS = \mathbb{R}^2$) changes to $\bs_{t+1} = \bs_t + \ba_t$.
The agent always starts on the left $\bs_0 = (-2, 0)$ and aims to go to its goal on the right $\bs_{\mathrm{fin}} = (2,0)$. However, the unit circle is a danger zone, in which the agent is subject to highly negative rewards (red shaded area).

We consider evaluation policies $\pi_y$ with an intermediate goal $\bs_{\mathrm{im}} = (0, y)$ on the y-axis that goes in a straight line from $\bs_0$ to $\bs_{\mathrm{im}}$ and then in a straight line from $\bs_{\mathrm{im}}$ to the goal $\bs_{\mathrm{fin}}$.
Note that policies $\pi_y$ with $|y| \leq 1.155$ 
are unsafe.

We generate an offline dataset by rolling out the behavior policy $\pi_{1.6}$ with Gaussian action noise with a standard deviation of $0.1$.
Then, we evaluate $\pi_{1.1}$, which is unsafe (see black trajectory), by rolling it out using \textsc{HAMBO-CA}.

We compare this to a neutral variant that predicts the next state with the predictive mean $\bmu_\Theta(\bs, \ba)$, i.e., without pessimism. As we can observe from the yellow trajectory, it falsely estimates $\pi_{1.1}$ as safe, that is, it predicts that the trajectory lies outside of the danger zone. 
The trajectories with the adversarial transition model and the corresponding epistemic confidence sets for every step are depicted in Fig~\ref{fig:toyexample}. The adversary successfully moves the prediction towards the danger zone within the confidence set, and, thus, correctly estimates the policy to be unsafe.
Overall, this demonstrates a failure case of (neutral) off-policy evaluation and shows how \textsc{HAMBO} reliably gives a conservative estimate of the policy value through its pessimistic transition model.
\vspace{-3mm}
\subsection{Empirical convergence of \textsc{HAMBO}} %
\vspacesubcaptionlow
\vspacesubcaption
\looseness=-1
For GP models, we show that \textsc{HAMBO} estimates converge to the true policy values (Theorem \ref{theorem:consistency}). Now, we empirically evaluate the behavior of GP-based \textsc{HAMBO} with an RBF kernel, as the number of offline data points grows. 
To this end, we consider two environments; a simple 2D PointEnv ($\calS=\mathbb{R}^2$, $\calA = [-1, 1]^2$), similar to the PointSaftey environment, and the Pendulum-v1 environment from the OpenAI Gym \citep{brockman2016openai}. 
In the PointEnv, the agent has to navigate the origin and accordingly receives the negative distance to the origin as a reward.

To generate the offline dataset, we collect transition data by uniformly sampling states and actions from the state and action space respectively. For the PointEnv, we restrict the sampled states to $[-40, 40]^2$ which covers the relevant part of the state space. 
As the evaluation policy, we use a proportional controller for the PointEnv, and a controller learned with SAC for the Pendulum. 

Figure~\ref{fig:gp_consistency} plots the \textsc{HAMBO} estimates $\tilde{J}(\pi_e)$ for a varying number of offline datapoints $n = |\calD_b|$. We notice that in the PointEnv, when we have insufficient data (here, ca. $n \leq 150$), the epistemic confidence regions of our GP model are large enough so that the transition model adversary sometimes manages to steer the policy outside the data support where the epistemic uncertainty is even higher. As a result, we see that $\tilde{J}(\pi_e)$ are initially far below the true expected return $J(\pi_e)$. However, as $n$ increases, the GP uncertainty regions become smaller, and, as we can observe in Figure~\ref{fig:gp_consistency}, $\tilde{J}(\pi_e)$ becomes an increasingly tighter lower bound, approaching $J(\pi_e)$ for both the environments. %

\begin{figure*}[t]
\centering
\includegraphics[width=0.32\textwidth]{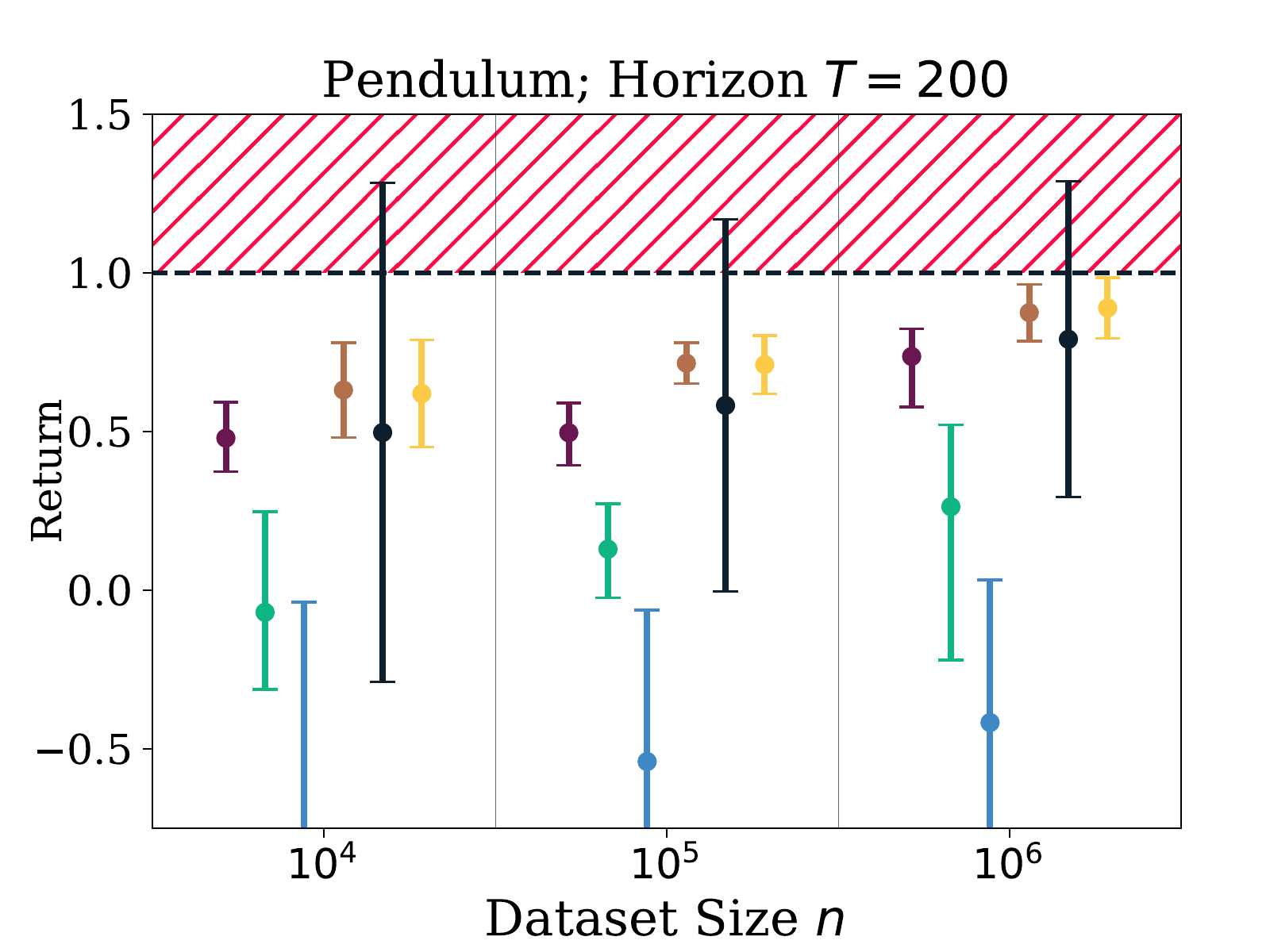}
\includegraphics[width=0.32\textwidth]{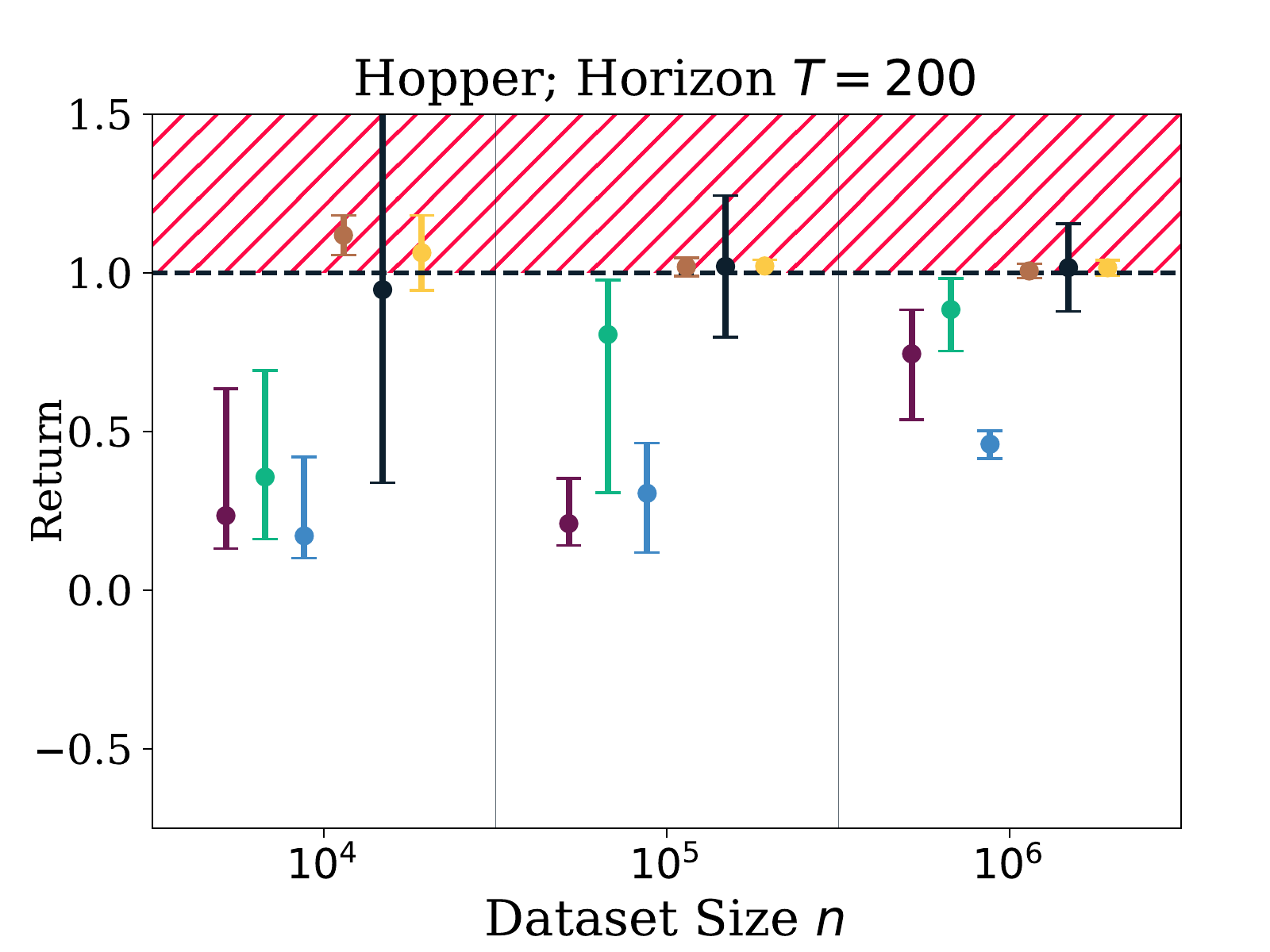}
\includegraphics[width=0.32\textwidth]{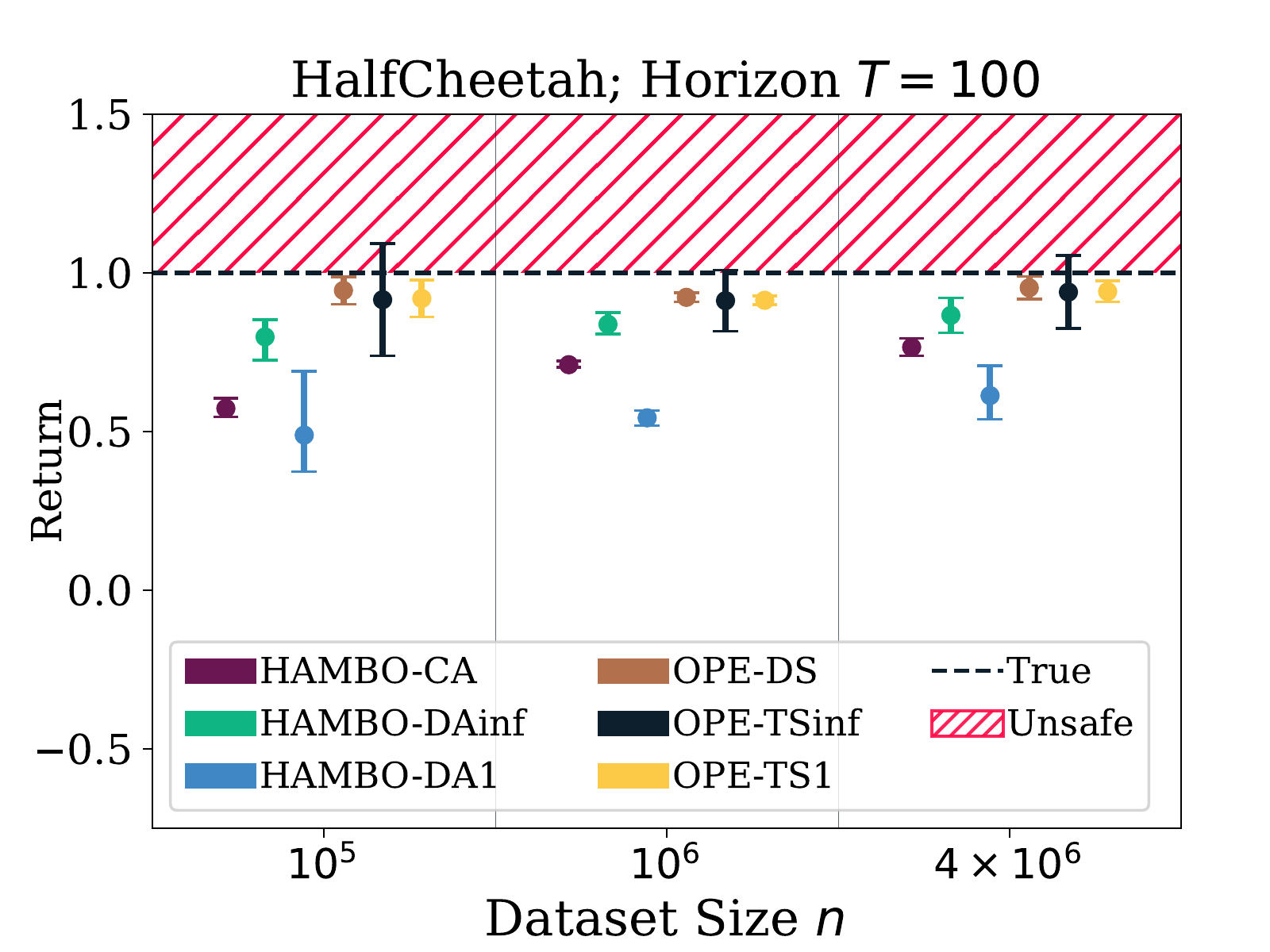}
\vspacecaption
\caption{\textsc{HAMBO} variants and neutral OPE baselines for continuous control. 
Unlike neutral OPE, which frequently overestimates the true expected return, \textsc{HAMBO} always yields a valid lower bound, which becomes more accurate with $n$.\looseness-1 \label{fig:ope_data}}
\vspacefigure
\end{figure*}
\begin{figure*}[!ht]
\centering
\includegraphics[width=0.32\textwidth]{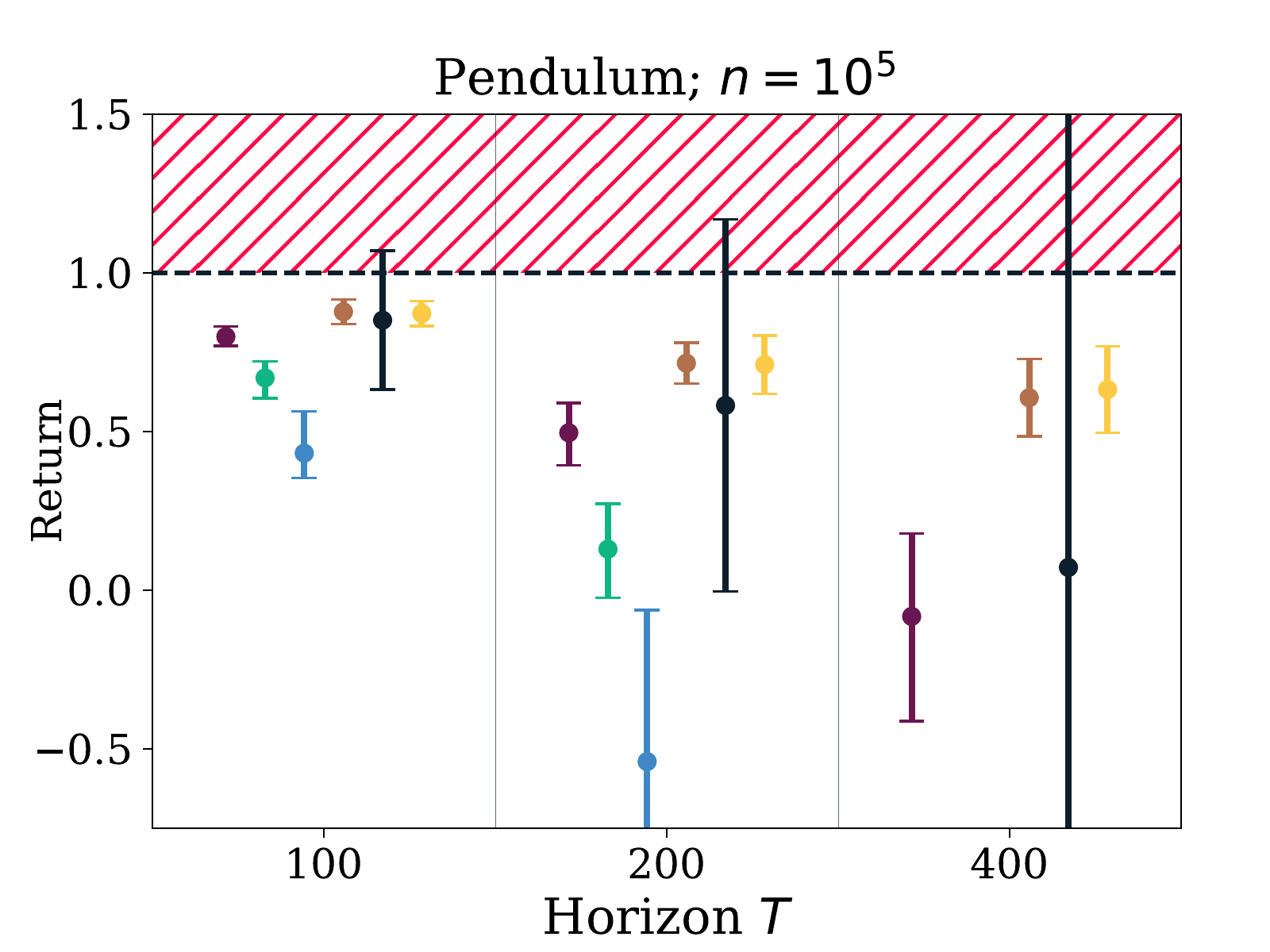}
\includegraphics[width=0.32\textwidth]{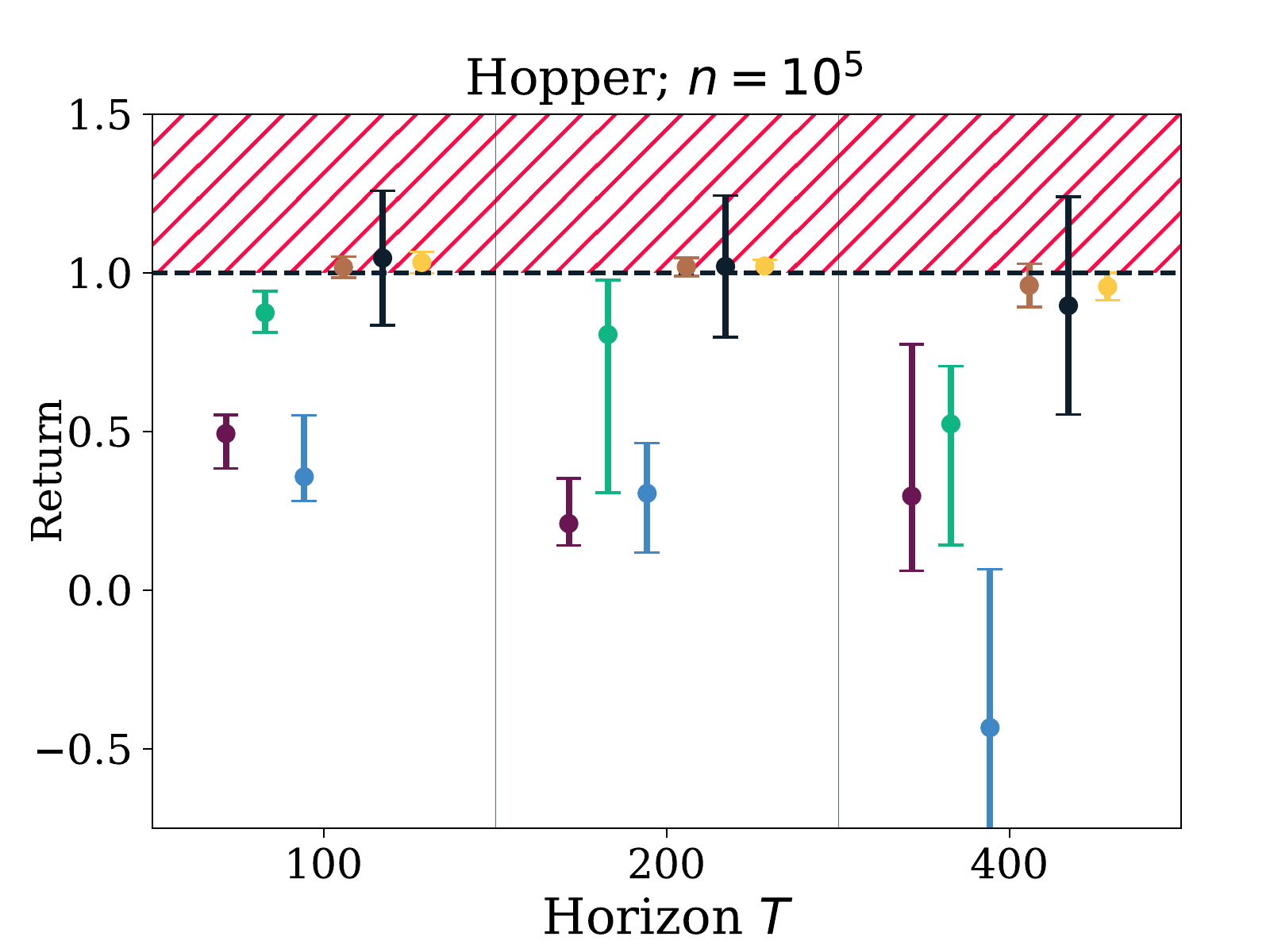}
\includegraphics[width=0.32\textwidth]{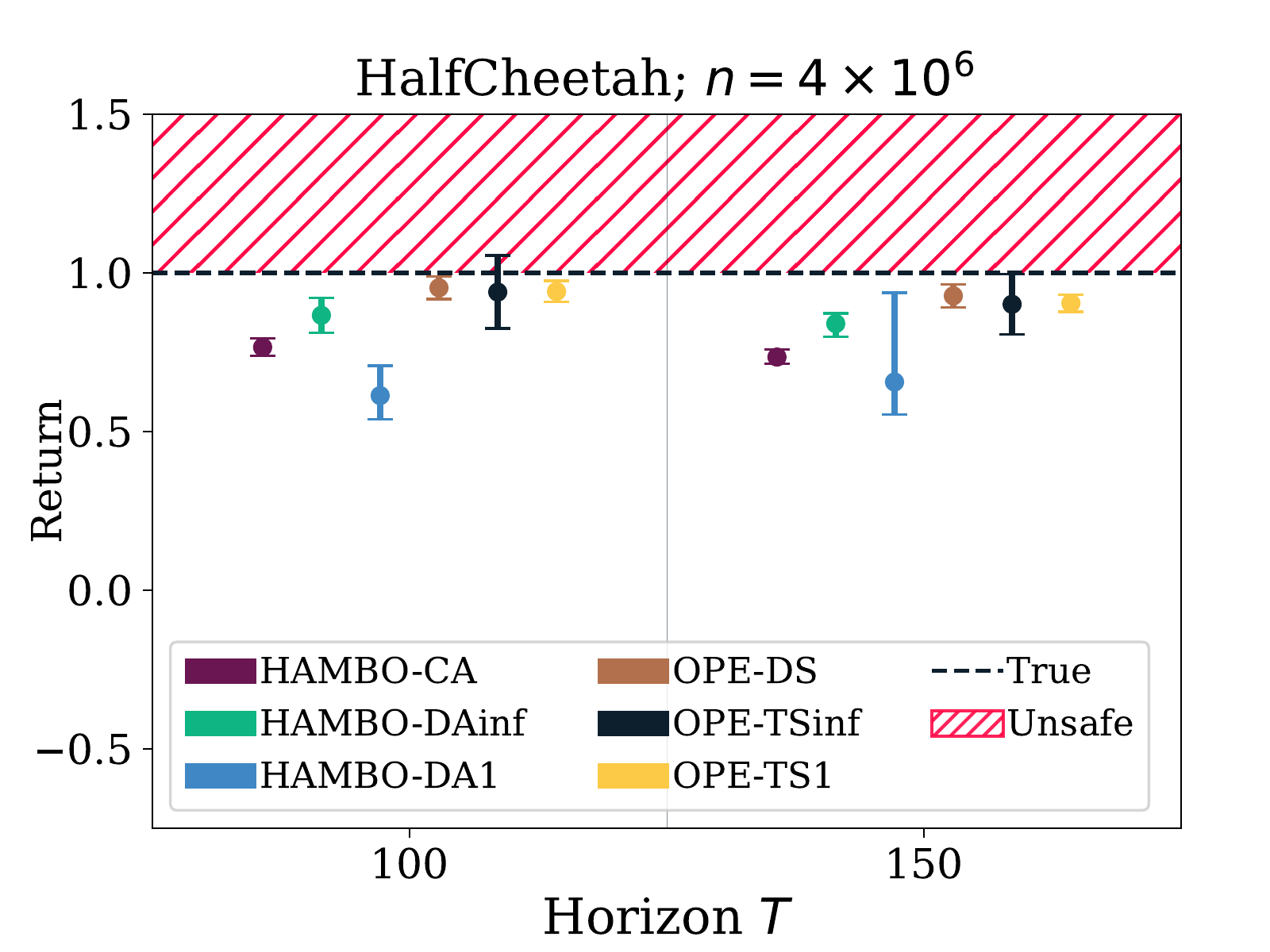}
\vspacecaption
\caption{\textsc{HAMBO} variants and neutral OPE baselines for different horizons. With longer horizons, the variance of the neutral OPE estimates increases and \textsc{HAMBO} lower bounds become looser.}
\label{fig:ope_horizon}
\vspacefigure
\end{figure*}
\vspacesubcaption
\vspace{-3mm}
\subsection{HAMBO for Continuous Control}
\vspacesubcaptionlow
\vspace{-1.5mm}
\looseness-1 We evaluate the NN-based \textsc{HAMBO} methods from Section \ref{sec:hambo_nn} on the continuous control tasks Pendulum-v1, Hopper-v3, and HalfCheetah-v3 from the OpenAI Gym and compare them to respective neutral (non-pessimistic) OPE methods.

\looseness -1 Our general methodology is as follows:
For a given environment, we first train a policy using the SAC algorithm \cite{haarnoja2018sac,haarnoja2018sac2} and save several checkpoints of the agent. Then, some of the mediocre-performing checkpoints are rolled out to generate an offline dataset. After that, a given policy (usually one of the best checkpoints) is evaluated with the NN-based \textsc{HAMBO} variants. 

We compare our approach to neutral OPE variants that do not use a pessimistic transition model \citep{fonteneau2013batch}. In particular, we consider various trajectory uncertainty propagation methods from \citet{chua2018pets}, employed in the context of OPE: First, we consider \textsc{OPE-DS}, where the transition model is approximated by a Gaussian $p(\bs'|\bs, \ba) = \calN(\bs' ;  \bmu_\Theta(\bs, \ba); \bsigma_\Theta^2(\bs, \ba))$, here the variance is the sum of the epistemic and aleatoric variance. Second, we consider \textsc{OPE-TS1} where the transition model is the mixture of predictive Gaussians in (\ref{eq:predictive_gmm}). This means that, in every step, one of the NN models is chosen uniformly at random to compute the next state distribution. Third, we consider \textsc{OPE-TSinf}, where, for every episode, we randomly commit to one of the $K$ NNs.

\vspace{-1mm}
\looseness-1 We investigate the following three aspects: 1) whether a method yields reliable lower bounds, 2) the effect of the offline dataset size, and 3) the curse of long horizons. Figure \ref{fig:ope_data} and \ref{fig:ope_horizon} report the estimated expected policy returns, averaged over 5 seeds, alongside the corresponding confidence intervals. %

\textbf{Reliable Lower Bounds. }
The \textsc{HAMBO} variants are designed to give reliable lower bounds on the true expected return. The results in Figure \ref{fig:ope_data} and \ref{fig:ope_horizon} empirically confirm that, across all seeds, all NN-based \textsc{HAMBO} variants reliably provide lower bounds on $J(\pi_e)$, and, thus, fulfill the COPE requirements from Definition \ref{definition:conservativeope}. In contrast, the neutral OPE variants which do not introduce pessimism w.r.t. the epistemic uncertainty of the transition model fail to do so. In many cases, they overestimate the true policy value, particularly in the Hopper environment. This demonstrates the importance of pessimism in model-based COPE and affirms the validity of \textsc{HAMBO}, even with BNN models, where calibration (Definition \ref{assumption:calibration}) cannot be formally proven.

\vspace{-1mm}
\textbf{Offline Dataset Size and Tightness.}
\looseness -1 The difference between \textsc{HAMBO} estimates $\tilde{J}(\pi_e)$ and the true expected reward $J(\pi_e)$ depends on the strength of the transition adversary, which is limited by the size of the epistemic confidence sets. As the size of the offline datasets $\calD_b$ increases, we can generally expect the epistemic uncertainty to shrink. Thus, the adversary $\boldeta$ becomes less powerful and the \textsc{HAMBO} estimates become an increasingly tight lower bound.

\vspace{-1mm}
In Figure \ref{fig:ope_data}, we empirically investigate this effect by varying the offline dataset size $n$. As we hypothesized, we can observe the general trend that the \textsc{HAMBO} estimates come close to the true policy value, as $n$ increases.
Moveover, we observe that the \textsc{HAMBO-DA1} estimates are always strictly smaller than those of the \textsc{HAMBO-DAinf} variant. This is expected, since in the \textsc{DA1} variant, the adversary can pick the worst-case NN transition model at every step while in the case of \textsc{DAinf} the adversary can only do so per trajectory, and, thus has less power. Since our experiment results indicate that the pessimism in \textsc{HAMBO-DAinf} is sufficient to obtain reliable lower bounds in practice, we conclude that \textsc{HAMBO-DAinf} is the preferred choice among the two. While \textsc{HAMBO-DAinf} performs better in Hopper and HalfCheetah, \textsc{HAMBO-CA} yields the tightest lower bounds in the Pendulum environment.

\vspace{-1mm}
\textbf{The Curse of the Long Horizons.}
Finally, we investigate the effect of the horizon length $T$ on our COPE estimates. Over the course of a trajectory, the transition model estimation errors can compound and lead to large discrepancies. This is a well-studied phenomenon in model-based RL \citep[e.g. see][]{janner2019mbpo}. In our case, this is reflected by the worst-case lower bound in Theorem \ref{theorem:lower_bound} which depends exponentially on $T$.

\vspace{-1mm}
\looseness -1  To evaluate the empirical effect of horizon length, we report the (C)OPE estimates for an offline dataset of size $n= 10^5$ across varying horizon lengths: $T$ = 100, 200 and 400 for the Pendulum and Hopper. For HalfCheetah, we only report horizon lengths of $T$ = 100 and 150. %
Figure~\ref{fig:ope_horizon} displays the corresponding results. For an increasing horizon length, the variance of the neutral variants increases and the lower bounds of the conservative \textsc{HAMBO} estimates become looser. However, the observed decline in tightness in Figure~\ref{fig:ope_horizon} is much less pronounced than the exponential decline of the worst-case bound in Theorem \ref{theorem:lower_bound}.

\vspace{-1mm}
\looseness -1  For large horizon lengths, it can happen that the hallucinated trajectory under the pessimistic transition model strives far outside the support of the offline data. In such cases, unlike neutral OPE methods, \textsc{HAMBO} will still provide lower bounds on the true expected return. However, these bounds can be very pessimistic. For instance, this can be observed in the case of Pendulum, where for $T=400$ the estimates of \textsc{HAMBO-DA1} and \textsc{HAMBO-DAinf} go out of the chart.
Making accurate long-horizon predictions is generally very hard. For instance, this is discussed extensively in the context of model-based RL in \citet{janner2019mbpo}. Often, a discount factor is used when computing returns to alleviate these issues. We highlight that we work with undiscounted returns and continuous state-action spaces, and, thus, operate in the most challenging setting for OPE.

\vspacecaptionlow

\section{Related Work}
\vspacecaptionlow

\label{section:related}
This work mainly contributes to the literature on off-policy evaluation for MDPs, which we divide to three categories.

\looseness -1 
\textbf{Model-Free OPE. } The key challenge in OPE is to the distribution shift between behavior and evaluated policy. A popular natural approach to correct the distribution mismatch is to use importance sampling (IS) ratios to re-weight the rewards collected by the behavior policy \citep{precup2000importancesampling, dudik2011doubly} or to adjust the recursive updates when estimating the values directly via the Bellman equation \citep{precup2001off, sutton2015importancesamplingstatemarginal, hallak2017importancesamplingstatemarginal}. Some work also combine both approaches to obtain a more favorable bias-variance trade-off \citep{jiang2015importancesamplingdr, thomas2016data}. Unlike \textsc{HAMBO}, these approaches are model-free, i.e., they do not learn a model of the state transitions. However, they suffer from three key disadvantages: First, they have notoriously high variance, especially if the evaluated policy differs a lot from the behavior policy \cite{levine2020tutorial}. Second, they require the support of the behavior occupancy measure $\rho^{\pi_b}$ to contain the support of $\rho^{\pi_e}$ which is often not the case. In contrast, \textsc{HAMBO} still provides valid estimates in this scenario. Third, to compute the importance ratios, they assume access to the distribution of behavior policy which is almost never the case in practical applications where data is often collected by human experts. \textsc{HAMBO} does not require access to the behavior policy and, thus, is much more broadly applicable.

A recent line of work \citep{nachum2019dualdice, nachum2019algaedice, zhang2020gendice, yang2020off} estimates the state occupancy correction ratios via a form of fixed point iteration, and does not require access to the behavior policy. However, the Bellmann-like fixed point iteration is not applicable to the finite horizon case that we study in this paper. In addition, due to the fixed point iteration, it is very hard to quantify the uncertainty or bound error that is associated with such OPE estimates, making them poorly suited to COPE.

\looseness -1 \textbf{Model-Based OPE. } 
This approach first learns the transition dynamics, to then simulate rollouts with the evaluation policy $\pi_e$ and thereby estimate the expected reward of $\pi_e$ \citep[e.g.,][]{fonteneau2013batch, hanna2017bootstrap, kostrikov2020bootstrap}. Due to error in predicting the transitions, the resulting OPE estimate may overestimate the policy's performance which is prohibited in safety-critical applications.
Our approach additionally simulates pessimistic trajectories using the model's epistemic uncertainty, to avoid overestimation.
Further, to the best of our knowledge, Theorem \ref{theorem:consistency} is the first consistency result for model-based OPE.

\textbf{COPE and High-Confidence OPE. }
\looseness -1 We study the problem of COPE which seeks a high-probability lower bound on the expected return. This is is closely related to estimating confidence bounds for OPE. \citet{thomas2015hcope} provide such confidence bounds for IS-based OPE estimates. However, due to the high variance of IS estimates, such bounds are often very loose \citep{levine2020tutorial}.
\citet{kallus2020double} and \citet{shi2021debiasedope} propose a model-free approach to give asymptotically normal confidence intervals for directly $J(\pi)$. 
Assuming that the $Q$-function resides in an RKHS, \citet{feng2020accountable} and \citet{ feng2021nonasymptotic} present rates of convergence, under theoretically unverified assumptions about the MDP.
 These model-free approaches only work for discounted, infinite-horizon MDPs, thus, are not generally applicable to our finite-horizon setting.

\citet{hanna2017bootstrap, kostrikov2020bootstrap} use model-based bootstrapping to construct confidence intervals for the OPE estimates. \citet{kostrikov2020bootstrap} prove the asymptotic correctness of the bootstrap confidence intervals only for finite state-action spaces. In contrast, we show the validity of our COPE estimates non-asymptotically for any $|\calD_b|$, and in continuous state-action spaces.
Alternatively, \citet{fontenau2009lower_bound} and \citet{paduraru2013off} employ a Lipschitz argument to obtain valid COPE estimates. Our derivation of the worst-case lower bound in Theorem~\ref{theorem:lower_bound} also uses Lipschitz continuity. However, the \textsc{HAMBO} estimate provide a tighter lower bound on the true policy value, as we use the local confidence intervals rather than the global Lipschitz constants to introduce pessimism. Furthermore, unlike the mentioned work, \textsc{HAMBO} does not require knowledge of the Lipschitz constant
and works with sub-Gaussian noise.

\vspacecaption
\vspace{-1mm}
\section{Conclusion}
\vspacecaption
\vspace{-1mm}
{HAMBO}, a novel approach for COPE that forms a pessimistic estimate of the expected return by hallucinating adversarial trajectories within the epistemic confidence regions of the estimated transition model. We formally prove the validity and consistency of the resulting COPE estimates. We propose various scalable NN-based variants of \textsc{HAMBO} and empirically demonstrate that they give reliable and tight lower bounds on the true expected return.

Importantly, our approach does not require access to the probability distribution of the behavior policy and gives reliable estimates, even when the support evaluation policy's occupancy measure is not contained in the offline data distribution. This makes \textsc{HAMBO} particularly relevant for safety-critical real-world applications, where the offline data is mostly collected by human experts and we need to make reliable decisions about whether a given policy is good enough to be deployed. 

HAMBO can be naturally combined with other offline reinforcement learning (ORL) algorithms to solve safety-critical ORL tasks. We leave this for future work to investigate.

\begin{acknowledgements} %
 This research was supported by the European Research Council (ERC) under the European Union's Horizon 2020 research and innovation program grant agreement no.\ 815943 and the Swiss National Science Foundation under NCCR Automation, grant agreement 51NF40 180545. Jonas Rothfuss was supported by an Apple Scholars in AI/ML fellowship. We thank Sebastian Curi for contributing to the initial idea for this project.
\end{acknowledgements}

\bibliography{refs}

\appendix
\onecolumn
\section{Algorithm and Experiment Details}\label{app:algos}
In the following, we provide algorithmic formalizations and implementation details of the \textsc{HAMBO} framework and its practical variants which were discussed in the main paper. 

\subsection{Generic HAMBO algorithm from Section \ref{subsec:general_hambo_approach}}
First, we formalize the general HAMBO framework from Section \ref{subsec:general_hambo_approach}:
\begin{algorithm}[ht]
\caption{\textsc{HAMBO} Framework}
\label{alg:hambo_basic}
\begin{algorithmic}
\Require Offline dataset $\calD_b$, evaluation policy $\pi_e$, reward function $r(\cdot, \cdot)$, Horizon $T$, initial state distribution $p_0(\bs_0)$
\State $(\bmu_n, \bsigma_n, \beta_n) \leftarrow \mathrm{TrainModel}(\calD_b)$ \Comment{Train statistical model with offline data}
\State $\tilde p_{\bm{\eta}}(\bs_{t+1}\vert \bs_t, \ba_t) \leftarrow  p_{\beps}\big(\bs_{t+1} -  \bmu_n(\bs_t, \ba_t)
    - \beta_n \boldeta(\bs_t, \ba_t)  \bsigma_n(\bs_t, \ba_t) \big)$
    \Comment{Set up adversarial transition model.}
\State $\tilde J(\pi) \leftarrow \min_{\bm{\eta}} \mathbb{E}_{\bs_0 \sim p_0}[\mathbb{E}_{p_{\bm{\eta}},\pi}[\sum_{t=0}^T r(\bs_{t}, \ba_{t})]]$  \Comment{Optimize adversary to get pessimistic value estimate.}\\
\Return  $\tilde J(\pi)$ %
\end{algorithmic}
\end{algorithm}

We estimate $J_{\tilde{p}_{\boldeta}}(\pi_e) = \mathbb{E}_{\bs_0 \sim p_0}[\mathbb{E}_{p_{\bm{\eta}},\pi}[\sum_{t=0}^T r(\bs_{t}, \ba_{t})]]$ via Monte Carlo estimation, i.e., we roll out $L$ trajectories and estimate the expectation as the average of the trajectory return:
\begin{equation} \label{eq:mc_estimation_return}
    \hat{J}_{\tilde{p}_{\boldeta}}(\pi_e) = \frac{1}{L} \sum_{l=1}^{L} \sum_{t=0}^T r(\bs_{l,t}, \ba_{l, t}) ~~ \text{where} ~~ \bs_{l,0} \sim p_0, ~ \ba_{l,t} \sim \pi(\ba| \bs_{l,t}), ~ \bs_{l,t+1} \sim  \tilde p_{\bm{\eta}}(\bs' \vert \bs_{l,t}, \ba_{l,t}) 
\end{equation}

The optimization of the advesary corresponds to a standard optimal control problem for which we use traditional methods such as trajectory optimization or model-free RL algorithms such as SAC.
\subsection{BNN Based HAMBO Variants}

\subsubsection{The BNN model}
We use fully connected neural networks with 4 hidden layers each of size 256 with ReLU activation functions.
Before training, the offline data inputs and targets are standardized. The NN takes the concatenated state and action as input (i.e., $d_s + d_a$ dimensional) and outputs a vector of size $2 d_s$ which is split into two vectors of size $d_s$. The first one corresponds to the mean prediction $\bh_{\btheta}(\bs, \ba)$ and the second one is the raw the aleatoric standard deviation which is fed through a softplus function to ensure positivity of $\bnu^2_{\btheta}(\bs, \ba)$

As BNN prior we use a standard Normal distribution over the NN parameters $\btheta$, i.e., $p(\btheta) = \calN(\btheta ; \boldsymbol{0}, \bm{I})$. However, as commonly done for BNNs to alleviate the problems of prior misspecification, we add a temperature parameter $\tau$ to the prior, so that we have $p(\btheta| \calD_b) \propto p(\calD_b | \btheta) p(\btheta)^\tau$. This hyper-parameter is chosen to as $\tau=0.0001$ for Pendulum and Hopper and $\tau=0.01$ for the HalfCheetah control environment.

We use Stein Variational Gradient Descent (SVGD) \citep{liu2016svgd} for approximate posterior inference. In particular, we approximate the posterior $p(\btheta| \calD_b)$ with $K$ NN particles $\{ \btheta_1, ..., \btheta_K\}$. After randomly initializing the parameters of the $K$ NNs, the parameters are iteratively updated with the SVGD update rule:
 \begin{equation}
     \btheta_k \leftarrow \btheta_k+  \frac{1}{K} \sum_{k'=1}^K \left [k(\btheta_{k'}, \btheta_k) \nabla_{\btheta_{k'}} \log p(\btheta_{k'}| \calD_b) +  \nabla_{\btheta_{k'}} k(\btheta_k', \btheta_k) \right] ~~ \forall k = 1, ..., K ~.
 \end{equation}
Here $k(\cdot, \cdot)$ is a kernel function on the space of NN parameters vectors. In our experiments, we use an RBF kernel $k(\btheta, \btheta') = \exp \left( - \norm{\btheta - \btheta'}^2 / (2 \ell) \right)$ with a length scale of $\ell = 10$ and $K=5$ NN particles. Note that the kernel here is different from the one in Section \ref{section:hambo_gp_models}. Algorithm \ref{alg:SVGD} summarizes how to obtain the SVGD BNN posterior approximation.

\begin{algorithm}[ht]
\caption{\textsc{SVGD}}
\label{alg:SVGD}
\begin{algorithmic}
\Require Training data $\calD$, number of particles $K$
\State Initialize NN parameter vectors $\btheta_1, ..., \btheta_K $ \Comment{Initialize SVGD particles.}
\While{not converged}
    \State $\log p(\btheta_{k}| \calD) \leftarrow \log p(\calD| \btheta_k) + \tau \log p(\btheta_k)  ~~\forall k = 1, ..., K$
    \State $\btheta_k \leftarrow \btheta_k+  \frac{1}{K} \sum_{k'=1}^K \left [k(\btheta_{k'}, \btheta_k) \nabla_{\btheta_{k'}} \log p(\btheta_{k'}| \calD) +  \nabla_{\btheta_{k'}} k(\btheta_k', \btheta_k) \right] ~~ \forall k = 1, ..., K$
\EndWhile \\
\Return $\{\bh_{\btheta_1}, \bnu^2_{\btheta_1}, \cdots,\bh_{\btheta_K}, \bnu^2_{\btheta_K}\} $ \Comment{Return the $K$ NN predictive mean and aleatoric variance functions}
\end{algorithmic}
\end{algorithm}

\subsection{Recalibration of the BNN uncertainty estimates} 
To obtain well-calibrated confidence sets for HAMBO, we recalibrate the BNNs predictive distribution. In particular, we use temperature scaling based on the regression calibration error \citet{kuleshov2018calibration}. We perform re-calibration based on the predictive distribution $\calN(\bmu_\Theta(\bs, \ba), \bsigma^2_{\Theta}(\bs, \ba))$. The calibration error compares the predictive quantiles of this Normal distribution with the corresponding empirical frequencies of data points, that fall below the predicted quantiles. Formally, we define $\Phi_{\boldsymbol{\tau}}^{-1}(\alpha; \bs, \ba): [0,1]^{d_s} \rightarrow \calS$ as the quantile function (inverse cumulative density function) of $\calN(\bmu_\Theta(\bs, \ba), \boldsymbol{\tau}^2 \bsigma^2_{\Theta}(\bs, \ba))$ where $\boldsymbol{\tau} \in \mathbb{R}^{d_s}$ is the temperature scaling vector. Given a calibration dataset $\calD_c = \{(\bs, \ba, \bs') \}$, the calibration error \citep{kuleshov2018calibration} for multivariate distributions follows as 
\begin{equation}
    \mathrm{CalErr}(\boldsymbol{\tau})  \coloneqq \frac{1}{d_s} \sum_{j=1}^{d_s} \frac{1}{|A|} \sum_{\alpha \in A}  \left(\mathrm{EmpFreq}(\alpha, \boldsymbol{\tau})_j - \alpha \right)^2 ~,
\end{equation}
where $A = \{0.1, \cdots, 0.9, 0.99\}$ is a set of confidence levels and
\begin{equation}
    \mathrm{EmpFreq}(\alpha; \boldsymbol{\tau}) \coloneqq \frac{1}{|\calD_c|} \sum_{(\bs, \ba, \bs') \in \calD_c} \mathbf{1}\{ \bs' \leq \Phi_{\boldsymbol{\tau}}^{-1}(\alpha; \bs, \ba) \}
\end{equation}
is a vector-valued function of the (per dimension) empirical frequencies of the prediction targets that fall below the $\alpha$ quantile.
Finally, we recalibrate the BNN predictions, by choosing the variance scaling vector $\boldsymbol{\tau}$ such that the calibration error is minimized, i.e., we choose
\begin{equation}
    \boldsymbol{\tau}^* = \argmin_{\boldsymbol{\tau}} \mathrm{CalErr}(\boldsymbol{\tau})  ~.
\end{equation}
Algorithm~\ref{alg:BNN_calibrate} summarizes this BNN re-calibration procedure:
\begin{algorithm}[ht]
\caption{\textsc{CalibrateBNN}}
\label{alg:BNN_calibrate}
\begin{algorithmic}
\Require calibration dataset $\calD_c$, predictive mean $\bmu(\cdot, \cdot)$, predictive variance $\bsigma^2(\cdot, \cdot)$
\State $A \leftarrow \{0.1, \cdots, 0.9, 0.99\}$. \Comment{Fix a set of confidence levels.}
\State $\Phi^{-1}(\cdot; \bs, \ba, \boldsymbol{\tau}): [0,1] \mapsto \sR^{d_s}$ as the inverse CDF of the Gaussian distribution $\calN(\bmu(\bs, \ba), \boldsymbol{\tau}^2\bsigma^2(\bx_i))$.
\State Define $\mathrm{EmpFreq}(\alpha; \boldsymbol{\tau}) \leftarrow \frac{1}{|\calD_c|} \sum_{(\bs, \ba, \bs') \in \calD_c} \mathbf{1}\{ \bs' \leq \Phi_{\boldsymbol{\tau}}^{-1}(\alpha; \bs, \ba, \boldsymbol{\tau}) \}$ 
\State Define $\mathrm{CalErr}(\boldsymbol{\tau})  \leftarrow \frac{1}{d_s} \sum_{j=1}^{d_s} \frac{1}{|A|} \sum_{\alpha \in A}  \left(\mathrm{EmpFreq}(\alpha, \boldsymbol{\tau})_j - \alpha \right)^2$\\
\Return $\arg\min_{\boldsymbol{\tau}} \mathrm{CalErr}(\boldsymbol{\tau}) $ \Comment{Choose $\boldsymbol{\tau}$ that minimizes the calibration error}
\end{algorithmic}
\end{algorithm}

\subsection{The NN-Based HAMBO variants}
Here, we provide algorithmic descriptions of the NN-Based HAMBO variants from Section \ref{sec:hambo_nn} as well as details about their implementation and how the corresponding experiments were conducted.

\subsubsection{HAMBO with a Continuous Adversary (HAMBO-CA)}

HAMBO-CA directly reflects the hallucinated adversarial transition model, introduced in Section \ref{section:hambo_general_idea}. The adversary $\boldeta(\bs, \ba) \in [-1, 1]^{d_s}$ chooses the mean of the Gaussian transition probability from the epistemic confidence set, i.e.,
\begin{align}
\tilde{p}_{\boldeta}(\bs'| \bs, \ba) \coloneqq \calN \big(\bs';  \bmu_\Theta(\bs, \ba) +  \tau^2 \boldeta(\bs, \ba)  \bsigma^2_{\Theta, e} , \bsigma^2_{\Theta, a}(\bs, \ba) \big) 
\end{align}
For obtaining the corresponding conservative value estimate $\tilde{J}(\pi_e) = \min_{\boldeta} J_{\tilde{p}_{\boldeta}}(\pi_e)$, we need to find the adversary $\boldeta^\star$ that minimizes the expected return. For this, we parameterize the adversary $\boldeta$ as a neural network policy with two hidden layers of size 256 with ReLU activations and a tanh squashed Gaussian conditional distribution over the adversary actions in $[-1, 1]^{d_s}$. We use SAC \citep{haarnoja2018sac2} to maximize the negative expected return of the adversary policy. As usual, to stabilize the SAC training and avoid Q-value overestimation, we use double critics and trailing target critics. The SAC training is conducted in rounds consisting of rollouts of 1000 episodes under the hallucinated transition model where actions are chosen by $\pi_e$, followed by 1000 gradient steps on the SAC objectives. For the gradient steps, we use a batch size of 1024 and the Adam optimizer with a learning rate of $10^{-3}$ for critic and policy and $5 * 10^{-5}$ for the SAC entropy parameter. After SAC has converged, we take the adversary policy $\boldeta^\star$ and estimate the expected return $\hat{J}_{\tilde p_{\boldeta^\star}}(\pi_e)$ of $\pi_e$ under the adversary transition model, induced by $\boldeta^\star$ with $L = 10^4$ trajectories (see Eq. \ref{eq:mc_estimation_return}). The HAMBO-CA method is summarized in Algorithm \ref{alg:hambo_ca}.

\begin{algorithm}[!h]
\caption{\textsc{HAMBO-CA}}
\label{alg:hambo_ca}
\begin{algorithmic}
\Require Offline dataset $\calD_b$, evaluation policy $\pi_e$, Number of BNN particles $K$
\State Select a subset of $\calD_b$ as calibration set $\calD_c$
\State $\{\bh_{\btheta_1}, \bnu^2_{\btheta_1}, \cdots,\bh_{\btheta_K}, \bnu^2_{\btheta_K}\} \leftarrow \mathrm{SVGD}(\calD_b \setminus \calD_c, K)$ \Comment{Train BNN via SVGD and get predictive NN functions}
\State $\bmu_\Theta(\bs, \ba) \leftarrow \frac{1}{K} \sum_{k=1}^K \bh_{\btheta_k}(\bs, \ba)$ \Comment{Calculate posterior mean.}
\State $\bsigma_{\Theta,e}^2(\bs, \ba) \leftarrow \frac{1}{K} \sum_{k=1}^K (\bh_{\btheta_k}(\bs, \ba) - \bmu_\Theta(\bs, \ba))^2$\Comment{Calculate epistemic uncertainty.}
\State $\bsigma_{\Theta,a}^2(\bs, \ba) \leftarrow\frac{1}{K}\sum_{k=1}^K \bnu^2_{\btheta_k}(\bs, \ba)$\Comment{Calculate aleatoric uncertainty.}
\State $\tau \leftarrow \mathrm{CalibrateBNN}(\calD_c, \bmu_\Theta, \bsigma^2_{\Theta, e}+\bsigma^2_{\Theta, a})$ \Comment{Calibrate the model}
\State Initialize adversary policy $\boldeta$
\State $\tilde{p}_{\boldeta}(\bs'| \bs, \ba) \leftarrow \calN \big(\bs' ; \bmu_\Theta(\bs, \ba) +  \tau^2 \boldeta(\bs, \ba)  \bsigma^2_{\Theta, e} , \bsigma^2_{\Theta, a}(\bs, \ba) \big)$ \Comment{Setup hallucinated adversarial transition model}
\State $\boldeta^\star \leftarrow \mathrm{SoftActorCritic}(-J_{\tilde p_{\boldeta}}(\pi_e), \boldeta)$ \Comment{Train adversary $\boldeta$ via SAC to maximize the negative return}
\State $\tilde{J}(\pi_e) \leftarrow \hat{J}_{\tilde p_{\boldeta^\star}}(\pi_e)$ \Comment{Estimate expected return of $\pi_e$ via sampling (see Eq. \ref{eq:mc_estimation_return})} \\
\Return $\tilde{J}(\pi_e)$
\end{algorithmic}
\end{algorithm}

\subsubsection{HAMBO with a Discrete Adversary (\textsc{HAMBO-DA1} and \textsc{HAMBO-DAinf})}
In the case of \textsc{HAMBO-DA1} the adversary $\vartheta$ has discrete action $\{1,..., K\}$, i.e. picking one of the $K$ particles. We parameterize the adversary policy as a neural network with two hidden layers of size 256 with ReLU activations and softmax-categorical distribution over the $K$ discrete actions. To train this adversary policy, we use clipped double DQN \citep{fujimoto2018td3}. The double DQN training is conducted in rounds consisting of rollouts of 1000 episodes under the hallucinated transition model where actions are chosen by $\pi_e$, followed by 1000 gradient steps on the DQN objectives. For the gradient steps, we use a batch size of 1024 and the Adam optimizer with a learning rate of $10^{-3}$. Once double DQN has converged, we take the adversary policy $\vartheta^\star$ and estimate the expected return $\hat{J}_{\tilde p_{\vartheta^\star}}(\pi_e)$ of $\pi_e$ under the adversary transition model, induced by $\vartheta^\star$ with $L = 10^4$ trajectories (see Eq. \ref{eq:mc_estimation_return}).
The overall \textsc{HAMBO-DA1} method is summarized in Algorithm \ref{alg:hambo_da-1}.
\begin{algorithm}[h]
\caption{\textsc{HAMBO-DA1}}
\label{alg:hambo_da-1}
\begin{algorithmic}
\Require Offline dataset $\calD_b$, evaluation policy $\pi_e$, Number of BNN particles $K$
\State  $\{\bh_{\btheta_1}, \bnu^2_{\btheta_1}, \cdots,\bh_{\btheta_K}, \bnu^2_{\btheta_K}\} \leftarrow \mathrm{SVGD}(\calD_b, K)$ \Comment{Train BNN via SVGD and get predictive NN functions}
\State Initialize adversary policy $\vartheta$
\State $\tilde{p}_{\vartheta}(\bs'| \bs, \ba) :=  \sum_{k=1}^K \vartheta(k | \bs, \ba) \calN \big(\bs' ;  \bh_{\btheta_k}(\bs, \ba) , \bnu^2_{\btheta_k}(\bs, \ba) \big)$ \Comment{Setup hallucinated adversarial transition model}
\State $\vartheta^\star \leftarrow \mathrm{DoubleDQN}(-J_{\tilde p_{\boldeta}}(\pi_e), \vartheta)$ 
\Comment{Train adversary $\vartheta$ via  to maximize the negative return}
\State $\tilde{J}(\pi_e) \leftarrow \hat{J}_{\tilde p_{\vartheta^\star}}(\pi_e)$ \Comment{Estimate expected return of $\pi_e$ via sampling (see Eq. \ref{eq:mc_estimation_return})} \\
\Return $\tilde{J}(\pi_e)$
\end{algorithmic}
\end{algorithm}

In contrast to \textsc{HAMBO-DA1}, \textsc{HAMBO-DAinf} uses a weaker adversary that has to commit to one of the BNN particles for the entire trajectory. As a result, the corresponding pessimistic HAMBO estimate can simply be chosen as the minimum of the expected evaluation policy return under each of the NN models in the particle approximation, i.e. $J(\pi_e) = \min_{k \in \{1, \dots, K\}} J_{p_{\btheta_k}}(\pi_e)$. The \textsc{HAMBO-DAinf} method is summarized in Algorithm \ref{alg:hambo_da-1}.

\begin{algorithm}[ht]
\caption{\textsc{HAMBO-DAinf}}
\label{alg:hambo_dainf}
\begin{algorithmic}
\Require  Offline dataset $\calD_b$, evaluation policy $\pi_e$, Number of BNN particles $K$
\State $\{\bh_{\btheta_1}, \bnu^2_{\btheta_1}, \cdots,\bh_{\btheta_K}, \bnu^2_{\btheta_K}\} \leftarrow \mathrm{SVGD}(\calD_b, K)$ \Comment{Train BNN via SVGD and get predictive NN functions}
\State $p_{\btheta_k}(\bs'| \bs, \ba) \leftarrow \calN \big(\bs' ;  \bh_{\btheta_k}(\bs, \ba) , \bnu^2_{\btheta_k}(\bs, \ba) \big)$ 
\State $\tilde J(\pi_e) \leftarrow \min_{k \in \{1, \dots, K\}} \hat{J}_{p_{\btheta_k}}(\pi_e)$ \Comment{Estimate return of $\pi_e$ for each model (see Eq. \ref{eq:mc_estimation_return})} and take minimum\\
\Return $\tilde J(\pi_e)$
\end{algorithmic}
\end{algorithm}

\newpage
\section{Proofs and Derivations}
\label{sec:missingproofs}

\begin{proof}[\textbf{Proof of Proposition~\ref{prop:valid_lower_bound}}]
By Assumption \ref{assumption:calibration} we have, with probability $1-\delta$, uniformly over $\calS \times \calA$, that
\begin{equation}
    | \bmu_n(\bs, \ba) - f(\bs, \ba) | \leq \beta_n(\delta) \bsigma_n(\bs, \ba) ~.
\end{equation}
Hence, there exists an (adversary) mapping $\boldeta^\dagger: \calS \times \calA \mapsto [-1, 1]^{d_s}$ such that every $\forall ~ \bs, \ba \in \calS \times \calA$ we have
\begin{equation}
f(\bs, \ba) = \bmu_n(\bs, \ba) + \beta_n(\delta) \boldeta^\dagger(\bs, \ba)  \bsigma_n(\bs, \ba) ~,
\end{equation}
and, thus the hallucinated transition model is equal to the true transition dynamics, i.e.,
\begin{align}
\tilde{p}_{\boldeta^\dagger}(\bs_{t+1}| \bs_t, \ba_t) = p_{\beps}\big(\bs_{t+1} - \bmu_n(\bs, \ba)
    + \beta \boldeta(\bs, \ba)^\dagger  \bsigma_n(\bs, \ba) \big) = p_{\beps}\big(\bs_{t+1} - f(\bs, \ba) \big) = p(\bs_{t+1}| \bs_t, \ba_t) ~.
\end{align}
Finally, we can use this to show
\begin{align}
    \tilde{J}(\pi_e) := \min_{\boldeta} J_{\tilde{p}_{\boldeta}}(\pi_e) \leq J_{\tilde{p}_{\boldeta^\dagger}}(\pi_e) =  J_{p}(\pi_e) = J(\pi_e) ~,
\end{align}
which concludes the proof.
\end{proof}

\begin{proof}[\textbf{Proof of Example \ref{example:reparametrizable_policy_is_w1_lip}}]
If $\pi(\ba|\bs)$ can be reparametrized as $g(\bs, \boldsymbol{\zeta})$, where $\boldsymbol{\zeta} \sim p(\boldsymbol{\zeta})$ and $g$ is $L_g$-Lipschitz, we have that the two random variables are equal in distribution, i.e.
\smash{
    $\ba \stackrel{d}{=} g(\bs, \boldsymbol{\zeta})$
}. Therefore,
\begin{align}
    \calW_1(\pi(\ba|\bs), \pi(\ba|\bs')) = &~\inf_{\gamma \in \Gamma(\pi(\ba|\bs), \pi(\ba|\bs'))} \E_{\ba, \ba' \sim \gamma} \left[ {\norm{\ba - \ba'}} \right] \\
    \leq &~ \E_{\ba, \ba' \sim \tilde{\gamma}} \left[ {\norm{\ba - \ba'}} \right] = \E_{\boldsymbol{\zeta}} \left[ {\norm{g(\bs, \boldsymbol{\zeta}) - g(\bs', \boldsymbol{\zeta})}} \right] \\
    \leq &~ L_g \norm{\bs - \bs'}
\end{align}
where $\tilde{\gamma}(\ba, \ba')$ is the joint probability distribution of $(g(\bs, \boldsymbol{\zeta}), g(\bs', \boldsymbol{\zeta}))$, and, thus a coupling. Hence, we have shown that $\pi(\ba, \bs)$ is $L_g$-Lipschitz w.r.t. the Wasserstein-1 distance.
\end{proof}

\subsection{Proof of Theorem~\ref{theorem:lower_bound}}

The following lemmata will be used to prove the theorem.

\begin{lemma}[Reparameterizability of two random variables with covariates]
\label{lemma:coupling_lemma}
Let $X$ and $Y$ be random variables with finite expectation and corresponding probability distributions $p$ and $q$. Then, we can reparameterize $Y$ as $Y \stackrel{d}{=} X + \bome_{|X}$, where $\bome_{|X}$ is a covariate that is generally dependent on $X$ and satisfies
\begin{equation}
    \E_X \E_{\bome | X} \left[ \norm{\bome} \right] = \calW_1(p, q) ~,
\end{equation}
where $\calW_1(p, q)$ is the Wasserstein-1 distance between $p$ and $q$.
\end{lemma}

\begin{proof}
Recall that the Wasserstein-1 distance is defined as infimum over couplings between $p$ and $q$, i.e.,
    \begin{equation} \label{eq:w1_def_23423}
        \calW_1 = \inf_{\gamma \in \Gamma(p, q)} \E_{X', Y' \sim \gamma} \left[ \norm{X'-Y'} \right]
    \end{equation}
    If the expectation of $p$ and $q$ is finite, then the infimum over couplings in (\ref{eq:w1_def_23423}) is attained for some $\gamma^*(x, y)$. 
    Now we construct the covariate $\bome_{|X}$ which is defined by applying the change of variable $g_x(x, y) \mapsto (x, y - x) = (x, \bome)$ to $\gamma^*$, so that we get $\tilde{\gamma}^*(x, \bome) = \gamma^*(x, x + \bome)$. The conditional distribution of the covariate $\bome_{|X}$ is 
    $$\tilde{\gamma}^*(\bome | x) = \frac{\gamma^*(x, x + \bome)}{\gamma^*(x)} = \frac{\gamma^*(x, x + \bome)}{p(x)}$$.

    Now, given our construction of $\bome_{|X}$, we aim to show that $Y \stackrel{d}{=} X + \bome_{|X}$. Define the random variable $Z:=X + \bome_{|X}$. Then we have 
    \begin{align}
        p(z) &= \int_{\calX} p(x, z-x) dx = \int_{\calX} p(x) \tilde{\gamma}^*(\underbrace{z-x}_{\bome}| x) dx \\
        &= \int_{\calX} \gamma^*(x, x+(z-x)) dx = \int_{\calX} \gamma^*(x, z) dx = q(z)
    \end{align}
    which shows that the pdf of $z$ is $q$, the probability density of $Y$. Since $\gamma^*(x, y)$ is the coupling that minimizes the transport cost, we can write
    \begin{align}
        \calW_1(p, q) = \inf_{\gamma \in \Gamma(p, q)} \E_{x', y' \sim \gamma} \left[ \norm{x'-y'} \right] 
        = &~ \E_{x', y' \sim \gamma^*} \left[ \norm{x'-y'} \right] = \E_{x' \sim p(x')} \E_{\bome \sim \tilde{\gamma}^*(\bome|x')} \left[ \norm{\bome} \right]
    \end{align}
which shows that $\E_X \E_{\bome | X} \left[ \norm{\bome} \right] = \calW_1(p, q)$, and, thus concludes the proof.
\end{proof}

\begin{corollary}
Let $\pi(\ba|\bs)$ be $L_\pi$-Lipschitz w.r.t. the Wasserstein-1 distance. For any arbitrary but fixed $\bs, \bs' \in \calS$ we denote $A$ and $A'$ as the random variables that follow the conditional distributions $\pi(\ba|\bs)$ and $\pi(\ba'|\bs')$ respectively. Then, we can construct a covariate $\bome_{|A}$ such that $A' \stackrel{d}{=} A + \bome_{|A}$ and
$\E_{A}\E_{\bome | A} \left[ \norm{\bome} \right] \leq L_\pi \norm{\bs - \bs'}$.
\label{corr:bound_on_coupling}
\end{corollary}
\begin{proof}
    The corollary directly follows from Lemma \ref{lemma:coupling_lemma} and the definition of the $L_\pi$-Lipschitz continuity w.r.t. the Wasserstein-1, i.e., that $\forall ~ \bs, \bs' \in \calS$ we have that $\calW_1\hspace{-2pt}\left(\pi(\ba|\bs), \pi(\ba|\bs') \right) \leq L_\pi \norm{\bs - \bs'}$.
\end{proof}

\begin{lemma}[Lipschitz continuity of Wasserstein-one distance implies Lipschitz continuity in expectation]
    Let $f: \mathcal{X}_1 \times \mathcal{X}_2 \rightarrow \mathcal{Y}$ be $L_f$ Lipschitz continuous and $x_2$ a random variable with distribution $p(\cdot|x_1)$ that is $L_p$ Lipschitz w.r.t. the Wasserstein-1 distance. Then we have
    \begin{equation*}
        \expvalue{x_2 \sim p(\cdot|x_1)}{f(x_1, x_2)} -    \expvalue{x'_2 \sim p(\cdot|x'_1)}{f(x'_1, x'_2)} \leq \Bar{L}_f||x_1 - x_1'||.
    \end{equation*}
    with $\Bar{L}_f = L_f(1+L_p)$.
    \label{lemma:lipschitz_composition}
\end{lemma}

\begin{proof}
    \begin{align*}
         \expvalue{x_2 \sim p(\cdot|x_1)}{f(x_1, x_2)} -    \expvalue{x'_2 \sim p(\cdot|x'_1)}{f(x'_1, x'_2)} &=  \expvalue{x_2 \sim p(\cdot|x_1)}{f(x_1, x_2)} - \expvalue{x_2 \sim p(\cdot|x_1)} {\expvalue{\bome \sim \Tilde{\gamma}^*(\bome|x_2)}{f(x'_1, x_2 + \bome)}} \tag{Lemma~\ref{lemma:coupling_lemma}. }\\
         &= \expvalue{x_2 \sim p(\cdot|x_1)} {\expvalue{\bome \sim \Tilde{\gamma}^*(\bome|x_2)}{{f(x_1, x_2)}  - f(x'_1, x_2 + \bome )}} \\
         &\leq L_f \expvalue{x_2 \sim p(\cdot|x_1)}{ \expvalue{\bome \sim \Tilde{\gamma}^*(\bome|x_2)}{
         {|| x_1 - x_1'||_2 + ||\bome||_2}}} \tag{Lipschitzness of $f$}\\
         &\leq L_f || x_1 - x_1'||_2 + L_fL_p||x_1 - x'_1||_2 \tag{Corollary~\ref{corr:bound_on_coupling}}\\
         &= L_f(1+L_p)||x_1 - x_1'||_2. %
    \end{align*}
\end{proof}

In the following, we bound the difference between the pessimistic and true return with the distance between the true and pessimistic trajectory using the Lipschitz continuity of the reward function and the policy's Wasserstein-one distance.
\begin{lemma}[Bound on difference between pessimistic and true return estimate]
\label{lemma:evaluation_error_bound}
Under Assumption \ref{assumption:lip_continuity} we have
\begin{align*}
\left|J(\pi_e) - \tilde{J}(\pi_e)\right|
&\leq \Bar{L}_r \expvalue{{\beps}_{0:T-1}, a_{0:T}}{\expvalue{{\bome}_{0:T}}{\sum_{t=0}^{T-1} ||\bs_t - \Tilde{\bs}_t||_2}}.
\end{align*}
where $\Bar{L}_r = L_r(1+L_{\pi})$.
\end{lemma}
\begin{proof}
We have
\begin{align*}
\left|J(\pi_e) - \tilde{J}(\pi_e)\right|&= \left|\expvalue{\bs_{0:T}, \ba_{0:T}}{\sum_{t=0}^{T-1}r(\bs_t, \ba_t)} - \expvalue{\tilde{\bs}_{0:T}, \tilde{\ba}_{0:T}}{\sum_{t=0}^{T-1}r(\tilde{\bs}_t,\tilde{\ba}_t)}\right|\\
&= \left|\expvalue{{\beps}_{0:T}, a_{0:T}}{\sum_{t=0}^{T-1}r(\bs_t, \ba_t)} - \expvalue{{\beps}_{0:T}, \ba_{0:T}} {\expvalue{\bome_{0:T}}{{\sum_{t=0}^{T-1}r(\tilde{\bs}_t,\ba_t + \bome_t)}}}\right| \tag{Lemma~\ref{lemma:coupling_lemma}}\\
&= \left|\expvalue{{\beps}_{0:T}, \ba_{0:T}}{\expvalue{\bome_{0:T}}{{\sum_{t=0}^{T-1}r(\bs_t, \ba_t) - r(\tilde{\bs}_t,\ba_t + \bome_t)}}}\right|.
\end{align*}
From Lemma~\ref{lemma:coupling_lemma}, know that $\mathbb{E}~\bome = \calW_1(\pi(\cdot \vert \bs_t), \pi(\cdot \vert \tilde{\bs}_t))$, where $\pi(\cdot \vert \cdot)$ is continuous w.r.t. the WD-1 distance. Therefore,
\begin{align*}
\left|J(\pi_e) - \tilde{J}(\pi_e)\right|&=\leq \expvalue{{\beps}_{0:T}, \ba_{0:T}}{\expvalue{\bome_{0:T}}{{\sum_{t=0}^{T-1} L_r(1+L_{\pi})||\bs_t - \Tilde{\bs}_t||_2}}} \tag{Lemma~\ref{lemma:lipschitz_composition}}\\
&= \Bar{L}_r \expvalue{{\beps}_{0:T}, \ba_{0:T}}{\expvalue{\bome_{0:T}}{{\sum_{t=0}^{T-1} ||\bs_t - \Tilde{\bs}_t||_2}}} \tag{$\Bar{L}_r = L_r(1+L_{\pi})$}.
\end{align*}
\end{proof}
Next, we bound the distance between the true and pessimistic trajectory with the epistemic uncertainty around the true trajectory.
\begin{lemma}[Bound on pessimistic and true trajectory]
\label{lemma:state_divergence_bound}
Under Assumption \ref{assumption:calibration} and \ref{assumption:lip_continuity} with probability at least $1-\delta$ for all $\boldeta: \mathcal{S} \rightarrow [-1,1]^{d_s}$ we have for all $t \in \{0,\dots,T\}$ that
\begin{align*}
\resizebox{0.99\linewidth}{!}{%
$\expvalue{{\beps}_{0:T}, \ba_{0:T}}{\expvalue{{\bome}_{0:T}}{\norm{\bs_{t+1} - \tilde \bs_{t+1}}}} 
\leq \left(1 + \sqrt{d_s}\right)\beta\sum_{i=0}^{(t+1)-1}\left(\Bar{L}_f + \left(1 + \sqrt{d_s}\right)\beta\Bar{L}_{\sigma}\right)^{(t+1)-1-i}\expvalue{{\beps}_{0:T}, \ba_{0:T}}{\expvalue{{\bome}_{0:T}}{\norm{\bsigma_n(\bs_i, \ba_i)}}}.$
}
\end{align*}
\end{lemma}
\begin{proof}
We prove by induction.
For $t=1$ we have 
\begin{align*}
\expvalue{{\beps}_{0:T}, \ba_{0:T}}{\expvalue{{\bome}_{0:T}}{\norm{\bs_1 - \tilde \bs_1}}} & =  \expvalue{{\beps}_{0:T}, \ba_{0:T}}{\expvalue{{\bome}_{0:T}}{\norm{f(\bs_0,\ba_0) + \beps_0 - \bmu_n(\bs_0,\ba_0) - \beta \bsigma_n(\bs_0,\ba_0)\boldeta(\bs_0, \ba_0) - \beps_0}}} \tag{Lemma~\ref{lemma:coupling_lemma}}\\
&\leq \expvalue{{\beps}_{0:T}, \ba_{0:T}}{\expvalue{{\bome}_{0:T}}{\norm{f(\bs_0,\ba_0) - \bmu_n(\bs_0, \ba_0)} + \norm{\beta_n \bsigma_n(\bs_0,a_0)\boldeta(\bs_0)}}}\\
&\leq \left(1 + \sqrt{d_s}\right)\beta_n\expvalue{{\beps}_{0:T}, \ba_{0:T}}{\expvalue{{\bome}_{0:T}}{\norm{\bsigma_n(\bs_0, \ba_0)}}} \tag{$\boldeta \in [-1,1]^{d_s}$}
\end{align*}

We get the induction hypothesis that for an arbitrary but fixed $t \geq 0$ we have
\begin{align*}
\resizebox{0.99\linewidth}{!}{%
$
\expvalue{{\beps}_{0:T}, \ba_{0:T}}{\expvalue{{\bome}_{0:T}}{\norm{\bs_t - \tilde \bs_t}}}
 \leq \left(1 + \sqrt{d_s}\right)\beta_n\sum_{i=0}^{t-1}\left(\Bar{L}_f + \left(1 + \sqrt{d_s}\right)\beta\Bar{L}_{\sigma}\right)^{t-1-i}\expvalue{{\beps}_{0:T}, \ba_{0:T}}{\expvalue{{\bome}_{0:T}}{\norm{\bsigma_n(\bs_i,\ba_i)}}}
 $
}
\end{align*}

Now for the induction step we can first derive
\begin{small}
\begin{align*}
\expvalue{{\beps}_{0:T}, \ba_{0:T}} {\expvalue{{\bome}_{0:T}}{\norm{\bs_{t+1} - \tilde \bs_{t+1}}}} &= \expvalue{{\beps}_{0:T}, \ba_{0:T}}{\expvalue{{\bome}_{0:T}}{\norm{f(\bs_t,\ba_t) + \beps_t - \bmu_n(\tilde \bs_t, \tilde \ba_t) - \beta_n \bsigma_n(\tilde \bs_t,\tilde \ba_t)\boldeta(\tilde \bs_t, \tilde \ba_t) - \beps_t}}} \\
&= \expvalue{{\beps}_{0:T}, \ba_{0:T}} {\expvalue{{\bome}_{0:T}}{\norm{f(\bs_t,\ba_t) - \bmu_n(\tilde \bs_t,\ba_t + \bome_t) - \beta_n \bsigma_n(\tilde \bs_t,\ba_t + \bome_t) \boldeta(\tilde s_t, \ba_t + \bome_t)}}} \tag{Lemma~\ref{lemma:coupling_lemma}}\\
&\leq \expvalue{{\beps}_{0:T}, \ba_{0:T}}{\expvalue{{\bome}_{0:T}}{\norm{f(\bs_t, \ba_t) - f(\tilde \bs_t,\ba_t + \bome_t) + f(\tilde \bs_t,\ba_t + \bome_t) - \bmu_n(\tilde \bs_t, \ba_t + \bome_t)}}}\\
&\quad \quad+ \expvalue{{\beps}_{0:T}, \ba_{0:T}}{\expvalue{{\bome}_{0:T}}{\norm{\beta_n \bsigma_n(\tilde \bs_t, \ba_t + \bome_t) \boldeta(\tilde s_t, \ba_t + \bome_t))}}}\\
&\leq \expvalue{{\beps}_{0:T}, \ba_{0:T}}{\expvalue{{\bome}_{0:T}}{\norm{f(\bs_t,\ba_t) - f(\tilde \bs_t,\ba_t + \bome_t)} + \norm{f(\tilde \bs_t,\ba_t + \bome_t) - \bmu_n(\tilde \bs_t, \ba_t + \bome_t)}}} \\
&\quad \quad+\expvalue{{\beps}_{0:T}, \ba_{0:T}}{\expvalue{{\bome}_{0:T}}{\norm{\beta_n \bsigma_n(\Tilde{\bs}_t, \ba_t + \bome_t)\boldeta(\tilde \bs_t, \ba_t + \bome_t)}}}\\
&\leq \expvalue{{\beps}_{0:T}, \ba_{0:T}}{\expvalue{{\bome}_{0:T}}{\Bar{L}_f\norm{\bs_t - \tilde \bs_t} + \left(1 + \sqrt{d_s}\right)\beta_n\norm{\bsigma_n(\tilde \bs_t, \ba_t + \bome_t)}}} \tag{Lemma~\ref{lemma:lipschitz_composition} and $\boldeta \in [-1,1]^{d_s}$}
\end{align*}
\end{small}
By applying the triangle inequality and adding and subtracting $\bsigma_n$ to the second term, 
\begin{small}
\begin{align*}
\expvalue{{\beps}_{0:T}, \ba_{0:T}}{\expvalue{{\bome}_{0:T}}{\norm{\bs_{t+1} - \tilde \bs_{t+1}}}} &\leq \expvalue{{\beps}_{0:T}, \ba_{0:T}}{\expvalue{{\bome}_{0:T}}{\Bar{L}_f\norm{\bs_t - \tilde \bs_t} + \left(1 + \sqrt{d_s}\right)\beta_n\norm{\bsigma_n(\Tilde{\bs}_t, \ba_t + \bome_t)}}}\\
&= \expvalue{{\beps}_{0:T}, \ba_{0:T}}{\expvalue{{\bome}_{0:T}}{\Bar{L}_f\norm{\bs_t - \tilde \bs_t}}} \\
&\quad\quad+ \expvalue{{\beps}_{0:T}, \ba_{0:T}}{\expvalue{{\bome}_{0:T}}{\left(1 + \sqrt{d_s}\right)\beta_n\norm{\bsigma_n(\tilde \bs_t, \ba_t+\bome_t) - \bsigma_n(\bs_t,\ba_t) + \bsigma_n(\bs_t,\ba_t)}}}\\
&\leq \expvalue{{\beps}_{0:T}, \ba_{0:T}}{\expvalue{{\bome}_{0:T}}{\Bar{L}_f\norm{\bs_t - \tilde \bs_t} + \left(1 + \sqrt{d_s}\right)\beta_n\left(\norm{\bsigma_n(\tilde \bs_t, \ba_t + \bome_t) - \bsigma_n(\bs_t, \ba_t)}\right)}} \\
&\quad\quad+ \left(1 + \sqrt{d_s}\right)\beta_n\expvalue{{\beps}_{0:T}, \ba_{0:T}}{\expvalue{{\bome}_{0:T}}{\left(\norm{\bsigma_n(\bs_t,\ba_t)}\right)}}\\
&\leq \expvalue{{\beps}_{0:T}, \ba_{0:T}}{\expvalue{{\bome}_{0:T}}{\Bar{L}_f\norm{\bs_t - \tilde \bs_t} + \left(1 + \sqrt{d_s}\right)\beta_n\left(\Bar{L}_{\sigma}\norm{\bs_t - \tilde \bs_t} + \norm{\bsigma_n(\bs_t,\ba_t)}\right)}} \tag{Lemma~\ref{lemma:coupling_lemma}}\\
&= \expvalue{{\beps}_{0:T}, \ba_{0:T}}{\expvalue{{\bome}_{0:T}}{\left(\Bar{L}_f + \left(1 + \sqrt{d_s}\right)\beta_n\Bar{L}_{\sigma}\right)\norm{\bs_t - \tilde \bs_t} + \left(1 + \sqrt{d_s}\right)\beta_n\norm{\bsigma_n(\bs_t,\ba_t)}}}
\end{align*}
\end{small}
Next, we apply the induction hypothesis
\begin{small}
\begin{align*}
\expvalue{{\beps}_{0:T}, \ba_{0:T}}{\expvalue{{\bome}_{0:T}}{\norm{\bs_{t+1} - \tilde \bs_{t+1}}}} &\leq \expvalue{{\beps}_{0:T}, \ba_{0:T}}{\expvalue{{\bome}_{0:T}}{\left(\Bar{L}_f + \left(1 + \sqrt{d_s}\right)\beta_n\Bar{L}_{\sigma}\right)\norm{\bs_t - \tilde \bs_t} + \left(1 + \sqrt{d_s}\right)\beta_n\norm{\bsigma_n(\bs_t, \ba_t)}}}\\
&\leq \left[\left(\Bar{L}_f + \left(1 + \sqrt{d_s}\right)\beta_n\Bar{L}_{\sigma}\right)\left(1 + \sqrt{d_s}\right)\beta_n\right] \\
& \quad\quad \times \Bigg( \sum_{i=0}^{t-1}\left(\Bar{L}_f + \left(1 + \sqrt{d_s}\right)\beta_n\Bar{L}_{\sigma}\right)^{t-1-i} \expvalue{{\beps}_{0:T}, \ba_{0:T}} {\expvalue{{\bome}_{0:T}}{\norm{\bsigma_n(\bs_i,\ba_i)}}} \\
&\quad\quad\qquad + \expvalue{{\beps}_{0:T}, \ba_{0:T}}{\expvalue{{\bome}_{0:T}}{{\norm{\bsigma_n(\bs_t, \ba_t)}}}}\Bigg)\\
&= \left(1 + \sqrt{d_s}\right)\beta_n\sum_{i=0}^{(t+1)-1}\left(\Bar{L}_f + \left(1 + \sqrt{d_s}\right)\beta_n\Bar{L}_{\sigma}\right)^{(t+1)-1-i}\expvalue{{\beps}_{0:T}, \ba_{0:T}}{\expvalue{{\bome}_{0:T}}{\norm{\bsigma_n(\bs_i, \ba_i)}}}\\
\end{align*} 
\end{small}
\end{proof}

Using the above lemmas, we present the proof to the main theorem.

\begin{proof}[\textbf{Proof of Theorem~\ref{theorem:lower_bound}}]
\begin{small}
    \begin{align*}
        \left|J(\pi_e) - \tilde{J}(\pi_e)\right| &\leq \Bar{L}_r \expvalue{{\beps}_{0:T}, \ba_{0:T}}{\expvalue{{\bome}_{0:T}}{\sum_{t=0}^{T-1} ||\bs_t - \Tilde{\bs}_t||_2}} \tag{Lemma~\ref{lemma:evaluation_error_bound}}\\
        &\leq \Bar{L}_r  \sum_{t=0}^{T-1} \left(1 + \sqrt{d_s}\right)\beta_n\sum_{i=0}^{(t+1)-1}\left(\Bar{L}_f + \left(1 + \sqrt{d_s}\right)\beta_n\Bar{L}_{\sigma}\right)^{(t+1)-1-i}\expvalue{{\beps}_{0:T}, \ba_{0:T}}{\expvalue{{\bome}_{0:T}}{\norm{\bsigma_n(\bs_i, \ba_i)}}} \tag{Lemma~\ref{lemma:state_divergence_bound}}
    \end{align*}
\end{small}
Since $\bar L_f \geq 1$, then for all $0\leq i\leq t$ and $0\leq t\leq T-1$, 
\[\left(\Bar{L}_f + \left(1 + \sqrt{d_s}\right)\beta_n\Bar{L}_{\sigma}\right)^{(t+1)-1-i} \leq \left(\Bar{L}_f + \left(1 + \sqrt{d_s}\right)\beta_n\Bar{L}_{\sigma}\right)^{T-1}  \]
which allows us to write, 
\begin{small}
\begin{align*}
\left|J(\pi_e) -  \tilde{J}(\pi_e)\right| &\leq \Bar{L}_r \left(1 + \sqrt{d_s}\right)\beta_n \left(1+\Bar{L}_f + \left(1 + \sqrt{d_s}\right)\beta_n\Bar{L}_{\sigma}\right)^{T-1} T \expvalue{{\beps}_{0:T}, \ba_{0:T}}{\expvalue{{\bome}_{0:T}}{{\sum_{t=0}^{T-1} \norm{\bsigma_n(\bs_t, \ba_t)}}}}\\
        &= \Bar{L}_r \left(1 + \sqrt{d_s}\right)\beta_n \left(1+\Bar{L}_f + \left(1 + \sqrt{d_s}\right)\beta_n\Bar{L}_{\sigma}\right)^{T-1} T \expvalue{{\bs}_{0:T}, \ba_{0:T}}{\sum_{t=0}^{T-1} \norm{\bsigma_n(\bs_t, \ba_t)}} \\
        &= \left[\Bar{L}_r \left(1 + \sqrt{d_s}\right)\beta_n \left(1+\Bar{L}_f + \left(1 + \sqrt{d_s}\right)\beta_n\Bar{L}_{\sigma}\right)^{T-1} T \right] \\
        &= \left[\Bar{L}_r \left(1 + \sqrt{d_s}\right)\beta_n \left(1+\Bar{L}_f + \left(1 + \sqrt{d_s}\right)\beta_n\Bar{L}_{\sigma}\right)^{T-1} T\right] \\
        &\hspace{2em} \times \left(\int_{\calS \times \calA} \sum_{t=0}^{T-1}  p(s_t=\bs, a_t=\ba | \pi_e, \mathcal{M})\norm{\bsigma_n(\bs, \ba)} d\bs d\ba \right)\\
        &=  \Bar{L}_r \left(1 + \sqrt{d_s}\right)\beta_n \left(1+\Bar{L}_f + \left(1 + \sqrt{d_s}\right)\beta_n\Bar{L}_{\sigma}\right)^{T-1} T \int_{\calS \times \calA} T \rho^{\pi_e}(\bs, \ba) \norm{\bsigma_n(\bs, \ba)} d\bs d\ba \tag{See Eq.~\ref{eq:state_occ_measure}}\\
        &= \Bar{L}_r \left(1 + \sqrt{d_s}\right)\beta_n \left(1+\Bar{L}_f + \left(1 + \sqrt{d_s}\right)\beta_n\Bar{L}_{\sigma}\right)^{T-1} T^2 \expvalue{\bs, \ba \sim \rho^{\pi_e}}{{\norm{\bsigma_n(\bs, \ba)}}}
    \end{align*}
    \end{small}
\end{proof}

In summary, the deviation between the true and pessimistic return is proportional to the expected epistemic uncertainty of the evaluation policy state-occupancy measure $\rho^{\pi_e}$, and the constant $C_n$ defined as
\begin{equation*}
    C_n \coloneqq \Bar{L}_r \left(1 + \sqrt{d_s}\right)\beta_n T^2\left(1+\Bar{L}_f + \left(1 + \sqrt{d_s}\right)\beta_n\Bar{L}_{\sigma}\right)^{T-1}.
\end{equation*}
In Appendix~\ref{app:consistency_proof}, we provide consistency guarantees for our method. In particular, we prove under further assumptions on the true dynamics function $f$, that $\left|J(\pi_e) - \tilde{J}(\pi_e)\right| \to 0$, for $n \to \infty$.

\subsection{Proof of known results for kernel methods} \label{app:gp_proofs}
We first recall the notion of maximum mutual information \citep{srinivas, elementsofIT}.
    The mutual information $I(\vx_{1:n}; k)$ quantifies the reduction in uncertainty due to the observations $\vx_{1:n}$. Given a GP model $\mathcal{GP}(0, k(\cdot, \cdot))$ and gaussian noise assumption, mutual information is equal to
\[
I(\vx_{1:n}) =  \frac{1}{2}\log\det(\mI + \sigma_\epsilon^{-2} \mK)
\]
with the kernel matrix $\mK = [k(\vx_i, \vx_j)]_{i,j \leq n}$.
    The maximum information capacity or maximum mutual information of a kernel $k$ is an upper bound on the mutual information, and is defined as 
    \begin{equation*}
        \gamma_n = \max_{\vx_{1:n}} I(\vx_{1:n}).
    \end{equation*}
  Table~\ref{tab:bound_on_gamma_n} shows the growth rate of $\gamma_n$ with $n$ for multiple different kernels.
  
\begin{proof}[Proof of Lemma \ref{lem:gp_calibrated}]
Let $\gamma_n$ be the maximum mutual information of $\mathcal{GP}(0, k(\cdot, \cdot))$. Set $\beta_n(\delta) := \left(B + \sigma_\epsilon \sqrt{2 (\gamma_n + 1 + \ln(d_s/\delta)} \right)$. 
     Element-wise application of Theorem 2 in \citet{chowdhury2017kernelized} over the dimensions of $\calS$ and taking a union bound proves the lemma.
\end{proof}

\begin{proof}[Proof of Lemma \ref{lemma:lip_f_sigma_gp}]
First, we prove the Lipschitz continuity of $\vf$.
By the Cauchy-Schwartz inequality, we have $\forall ~ \bx, \bx' \in \calX$
\begin{equation}
    |f_j(\bx) - f_j(\bx')| = | \langle f_j, k(\bx, \cdot) - k(\bx', \cdot) \rangle_k | \leq \normAny{f_j}_k ~ d_k(\bx, \bx') 
\end{equation}
Since $\normAny{f_j}_k \leq B, \forall j=1, ..., d_s$ and $d_k(\bx, \bx')$ is $L_k$-Lipschitz, we have that
\begin{equation}
\norm{\vf(\bx) - \vf(\bx')} = \sqrt{\sum_{j=1}^{d_s} (f_j(\bx) - f_j(\bx'))^2} \leq \sqrt{d_s B^2 d_k(\bx, \bx')^2} = \sqrt{d_s} B d_k(\bx, \bx') \leq \sqrt{d_s} B L_k \norm{\bx - \bx'} ;.
\end{equation}
Next, we show the Lipschitz continuity of the GP standard deviation. By Lemma 12 in \citet{curi2020hucrl}, we have, independent of $n$, $|\bsigma_n(\bx) - \bsigma_n(\bx)| \leq d_k(\bx, \bx')$ for the GP standard deviation. Now, we make a similar argument as above:
\begin{equation}
    \norm{\bsigma_n(\bx) - \bsigma_n(\bx)} \leq \sqrt{d_s d_k^2(\bx, \bx')} \leq  \sqrt{d_s} L_k \norm{\bx - \bx'}
\end{equation}
which shows that $\bsigma_n(\cdot)$ is $\sqrt{d_s} L_k$-Lipschitz.
\end{proof}

\subsection{Proof of Theorem~\ref{theorem:consistency}} \label{app:consistency_proof}
For showing consistency of our lower bound in Theorem~\ref{theorem:lower_bound} for the GP case, we first prove that the uncertainty with respect to an i.i.d., data sampling distribution $p(x)$ shrinks in expectation. 
\begin{lemma}[Shrinking uncertainty in expectation]
        Let $p(x)$ denote a data sampling distribution with compact support. Then the following holds for sequences $\{x_i\}^{n-1}_{i=0}$ sampled i.i.d.\ from  $p(x)$, 
          \begin{equation}
        C^2_{n}\expvalue{\vx_{1:n}\sim p}{\sigma^2(x_n|\{x_i\}^{n-1}_{i=0})} \leq C^2_{n}\expvalue{x_{1:n-1}\sim p}{\sigma^2(x_{n-1}|\{x_i\}^{n-2}_{i=0})}.
    \end{equation}
    \label{lemm:shrinking_var_in_exp}
\end{lemma}
\begin{proof}
    \begin{align*}
        C^2_{n}\expvalue{\vx_{1:n}\sim p}{\sigma^2(x_n|\{x_i\}^{n-1}_{i=0})} &=  C^2_{n}\expvalue{x_{1:n-1}\sim p}{\expvalue{x_{n}\sim p}{\sigma^2(x_n|\{x_i\}^{n-1}_{i=0})}} \\
        &= C^2_{n}\expvalue{x_{1:n-1}\sim p}{\expvalue{x\sim p}{\sigma^2(x|\{x_i\}^{n-1}_{i=0})}} \\
        &\leq C^2_{n}\expvalue{x_{1:n-1}\sim p}{\expvalue{x\sim p}{\sigma^2(x|\{x_i\}^{n-2}_{i=0})}} \tag{Monotonicity of variance}\\
        & =C^2_{n}\expvalue{x_{1:n-2}\sim p}{{\expvalue{x_{n-1}\sim p}{\expvalue{x\sim p}{\sigma^2(x|\{x_i\}^{n-2}_{i=0})\vert x_{n-1}}}}} \\
        &=C^2_{n}\expvalue{x_{1:n-2}\sim p}{\expvalue{x\sim p}{\sigma^2(x|\{x_i\}^{n-2}_{i=0})}} \tag{All points are sampled i.i.d from $p$} \\
                &=C^2_{n}\expvalue{x_{1:n-2}\sim p}{\expvalue{x_{n-1}\sim p}{\sigma^2(x_{n-1}|\{x_i\}^{n-2}_{i=0})}} \\
        &= C^2_{n}\expvalue{x_{1:n-1}\sim p}{\sigma^2(x_{n-1}|\{x_i\}^{n-2}_{i=0})} .
    \end{align*}
\end{proof}

\begin{lemma}[Bound on expectation of uncertainty at $n$]
    Let $p(\vx)$ denote the data sampling distribution with a compact support. Then the following holds for sequences $\{\vx_i\}^{n-1}_{i=0}$ sampled i.i.d.\ from  $p(\vx)$, 
    \begin{equation}
        n C^2_{n}\expvalue{\vx_{1:n}\sim p}{\sigma^2(\vx_n|\{\vx_i\}^{n-1}_{i=0})} \leq C^2_{n}\expvalue{\vx_{1:n}\sim p}{\sum^n_{j=1}\sigma^2(\vx_{j}|\{\vx_i\}^{j-1}_{i=0})}.
        \label{eq:variance_bound_with_n}
    \end{equation}
    Moreover, we have
    \begin{equation*}
        C^2_{n}\expvalue{\vx_{1:n}\sim p}{\sigma^2(\vx_n|\{\vx_i\}^{n-1}_{i=0})}  \leq \frac{C^2_n\gamma_n}{n},
    \end{equation*}
    where $\gamma_n$ represents the maximum information gain (\cite{srinivas, elementsofIT}).
    \label{lemm:bound_var_with_gamma}
\end{lemma}

\begin{proof}
    We prove by induction. For $n=1$, Eq.~\ref{eq:variance_bound_with_n} holds trivially.
    Now assume $n>1$,
    \begin{align*}
        n C^2_{n}\expvalue{\vx_{1:n}\sim p}{\sigma^2(\vx_n|\{\vx_i\}^{n-1}_{i=0})} &= C^2_{n}\expvalue{\vx_{1:n}\sim p}{\sigma^2(\vx_n|\{\vx_i\}^{n-1}_{i=0})} + (n -1 ) C^2_{n}\expvalue{\vx_{1:n}\sim p}{\sigma^2(\vx_n|\{\vx_i\}^{n-1}_{i=0})} \\
        &\leq C^2_{n}\expvalue{\vx_{1:n}\sim p}{\sigma^2(\vx_n|\{\vx_i\}^{n-1}_{i=0})} + (n-1) C^2_{n}\expvalue{\vx_{1:n-1}\sim p}{\sigma^2(\vx_{n-1}|\{\vx_i\}^{n-2}_{i=0})} \tag{Lemma~\ref{lemm:shrinking_var_in_exp}} \\
        &\leq C^2_{n}\expvalue{\vx_{1:n}\sim p}{\sigma^2(\vx_n|\{\vx_i\}^{n-1}_{i=0})} + C^2_{n}\expvalue{\vx_{1:n-1}\sim p}{\sum^{n-1}_{j=1}\sigma^2(\vx_{j}|\{\vx_i\}^{j-1}_{i=0})} \tag{By induction hypothesis} \\
        &= C^2_{n}\expvalue{\vx_{1:n}\sim p}{\sum^n_{j=1}\sigma^2(\vx_{j}|\{\vx_i\}^{j-1}_{i=0})}.
    \end{align*}

    Note, $\sum^{n}_{j=1} {\sigma^2(\vx_{j+1}|\{\vx_i\}^{j}_{i=0})}$ is a measure of the mutual information associated to the sampling scheme, and lower bounds the mutual information. 
    The mutual information $I(\vx_{1:n})$ quantifies the reduction in uncertainty due to the observations $\vx_{1:n}$~\cite{elementsofIT}. When $f \in \calH_k$, mutual information is equal to
\[
I(\vx_{1:n}) =  \frac{1}{2}\log\det(\mI + \lambda^{-1} \mK)
\]
with the kernel matrix $\mK = [k(\vx_i, \vx_j)]_{i,j \leq n}$.
    Moreover,
    \begin{equation*}
        \expvalue{\vx_{1:n}\sim p} {\sum^{n}_{j=1} {\sigma^2(x_{j+1}|\{x_i\}^{j}_{i=0})} } \leq I(\vx_{1:n}).
    \end{equation*}
    The maximum information gain, is an upper bound on the mutual information, and is defined as 
    \begin{equation*}
        \gamma_n = \max_{\vx_{1:n}} I(\vx_{1:n}).
    \end{equation*}
    Therefore, by definition of $\gamma_n$, it is greater
    than the mutual information of all sampling schemes within the the support of $p(x)$. 
        \begin{equation*}
        C^2_{n}\expvalue{\vx_{1:n}\sim p}{\sigma^2(x_n|\{x_i\}^{n-1}_{i=0})} \leq \frac{C^2_{n}\gamma_n}{n}
    \end{equation*}
    \cite{srinivas} derive the bounds on $\gamma_n$ (see Table~\ref{tab:bound_on_gamma_n}) for linear, RBF, and Matèrn kernels on compact and convex sets. Hence, we obtain that for the linear and RBF kernel, $C^2_n\gamma_n$ grows sublinearly in $n$, i.e., $\sfrac{C^2_n \gamma_n}{n} \to 0$ for $n\to \infty$. 
    \begin{table}[ht]
    \centering
    \begin{tabular}{c|c}
    Kernel                                   & Bounds on $\gamma_n$ for $x \in \mathbb{R}^d$ \\
    \hline
    Linear                                   &     $\mathcal{O}(d\log{n})$           \\
    RBF                                      &       $\mathcal{O}((\log{n})^{d+1})$         \\
    Matèrn $\nu>1/2$ &       $\mathcal{O}(n^{\frac{d}{2\nu + d}}\log^{\frac{2\nu}{2\nu+d}}(n))$         
    \end{tabular}
    \caption{Bounds on $\gamma_n$ from~\cite[Theorem 5.]{vakili2021information}}
    \label{tab:bound_on_gamma_n}
    \end{table}
\end{proof}
\begin{lemma}
    Let $p(\vx)$ denote a distribution with compact support, and assume that $\vx_1, \dots, \vx_{n-1}$ are i.i.d. samples from $p$. Then, the following holds,
    \begin{equation*}
      \sP\left(  \expvalue{\vx \sim p}{C_{n}\sigma_{n}(\vx)} = \calO\left((1+1/\sqrt{\delta})\sqrt{\frac{\gamma^{T+1}_n}{n}}\right),\, \forall \vx_{1:n-1}\right) \geq 1-\delta, \quad\quad \forall n \in \mathbb{N}.
    \end{equation*}
    For kernels with a maximum information capacity $\gamma_n = \calO(\text{poly}(\log(n))$ that grows at most polylogarithmically with $n$, we have that
    \begin{equation*}
        \mathbb{P}\left(\expvalue{x \sim p}{C_{n}\sigma_{n}(x)} \to 0 \text{ for } n \to \infty\right) = 1.
    \end{equation*}
    \label{lemma:consistency_of_behavioural_policy}
\end{lemma}
\begin{proof}
    From Lemma~\ref{lemm:bound_var_with_gamma}
    \begin{equation}
          \expvalue{x_{1:n-1} \sim p}{ C^2_{n}\expvalue{x \sim p}{\sigma^2_{n}(x)} }  \leq \frac{C^2_{n}\gamma_{n}}{n}.
    \end{equation}
    Using the Markov inequality, we get
     \begin{equation*}
        \mathbb{P}(X \geq a) \leq \frac{\E\left[X\right]}{a}.
    \end{equation*}
    Let $X$ denote $C^2_{n}\expvalue{x \sim p}{\sigma^2_{n}(x)}$. Then we have for all $a>0$, 
    \begin{equation*}
        \mathbb{P}(C^2_{n}\expvalue{x \sim p}{\sigma^2_{n}(x)} \geq a) \leq \frac{\expvalue{x_{1:n-1} \sim p}{ C^2_{n}\expvalue{x \sim p}{\sigma^2_{n}(x)} }}{a} \leq \frac{C^2_{n}\gamma_{n}}{na}.
    \end{equation*}
    Therefore, for $n\to \infty$, $C^2_{n}\expvalue{x \sim p}{\sigma^2_{n}(x)} \to 0$ almost surely if $\frac{C^2_{n}\gamma_{n}}{n} \to 0$ for $n\to \infty$. Now by definition of $C_n$ (Theorem~\ref{theorem:lower_bound}) and plugging in the choice of $\beta_n$ (Lemma~\ref{lem:gp_calibrated}), we have $\frac{C^2_n\gamma_n}{n} \propto \frac{\gamma^{T+1}_n}{n}$. By assumption, we have that $\gamma_n = \calO(\text{poly}(\log(n))$, and, thus $\frac{\gamma^{T+1}_n}{n} = \calO(\text{poly}(\log(n))^{T+1} / n) =\calO(\text{poly}(\log(n)) / n)$. Hence, $\frac{C^2_n\gamma_n}{n} \to 0$ for $n\to \infty$.
    For example, for the linear and RBF kernel, we have (see Table~\ref{tab:bound_on_gamma_n})
    \begin{align*}
        \frac{\gamma^{T+1}_n}{n} &= \calO\left(d^{T+1}\frac{\left(\log{n}\right)^{T+1}}{n}\right) \tag{Linear kernel}\\
          \frac{\gamma^{T+1}_n}{n} &= \calO\left(\frac{\left(\log{n}\right)^{(d+1)(T+1)}}{n}\right) \tag{RBF kernel}.
    \end{align*}

    Now to recover the rate of convergence, let $v = \expvalue{x \sim p}{C_{n}\sigma_{n}(x)}$. We study its variance and expectation with respect to $x_{1:n-1} \sim p$ for a fixed $n$. 
    We have \[
    \text{Var}[v] \leq  \expvalue{}{v^2} \leq  \expvalue{x_{1:n-1} \sim p}{ C^2_{n}\expvalue{x \sim p}{\sigma^2_{n}(x)} }\leq \frac{C^2_{n}\gamma_{n}}{n}.\]
    Additionally, $\expvalue{}{v} \leq \sqrt{\expvalue{}{v^2}} \leq C_{n}\sqrt{\frac{\gamma_{n}}{n}}$. Now, we apply the Chebyshev inequality, i.e., 
    \begin{equation*}
        \mathbb{P}(|v - \expvalue{v}{v}| \geq a) \leq \frac{\text{Var}[v]}{a^2} \leq \frac{\expvalue{v}{v^2}}{a^2}.
    \end{equation*}
Therefore, for $a^2 = \frac{\expvalue{v}{v^2}}{ \delta}$, we have with probability at least $1-{\delta}$, 
\begin{align*}
    v &\leq \expvalue{v}{v} + a \\
    &= \expvalue{v}{v} + \sqrt{\frac{\expvalue{v}{v^2}}{{ \delta}}}  \\
    &\leq \left(1 + \frac{1}{\sqrt{{\delta}}}\right) \sqrt{\expvalue{v}{v^2}}.
\end{align*}
Next, we plug in the definition of $v$, to get
\begin{align*}
        \expvalue{x \sim p}{C_{n}\sigma_{n}(x)} &\leq \left(1 + \frac{1}{\sqrt{{\delta}}}\right)\sqrt{\expvalue{x_{1:n-1} \sim p}{\expvalue{x \sim p}{ C^2_{n}\sigma^2_{n}(x)}}} \\
        &\leq \left(1 + \frac{1}{\sqrt{{ \delta}}}\right)C_n\sqrt{\frac{\gamma_{n}}{n}} = \calO\left(\left(1 + 1/\sqrt{\delta}\right)\sqrt{\frac{\gamma^{T+1}_n}{n}}\right).
\end{align*}
with probability at least $1-{ \delta}$.
\end{proof}

\begin{proof}[Proof of Theorem~\ref{theorem:consistency} (Consistency of HAMBO)]
    For the GP case we prove that the well calibration assumption, and the Lipschitz continuity of $f$ and $\sigma$ are satisfied (see Lemmas~\ref{lem:gp_calibrated}~and~\ref{lemma:lip_f_sigma_gp}). 
    This allows us to apply Theorem~\ref{theorem:lower_bound} and Proposition~\ref{prop:valid_lower_bound}, which gives with probability at least $1-\delta$ that,
        \begin{equation}\label{eq:theorem5_7}
       J(\pi_e) \geq \tilde{J}(\pi_e) \geq J(\pi_e) -  C_n ~ \expvalue{\bs, \ba \sim \rho^{\pi_e}}{ \norm{\bsigma_n(\bs, \ba)}}.
    \end{equation}
     To prove consistency, we then only need to show that $C_n ~ \expvalue{\bs, \ba \sim \rho^{\pi_e}}{ \norm{\bsigma_n(\bs, \ba)}}$ goes to $0$ for $n \to \infty$. 
    Since the support of the behavioural policy's state-occupancy measure $\rho^{\pi_b}$ is compact, and $\supp{\rho^{\pi_e}} \subseteq \supp{\rho^{\pi_b}}$, we have $\rho^{\pi_b}(s, a) \geq \hat{C}\rho^{\pi_e}(s,a)$ for all $(s,a) \in \calS \times \calA$, and some $\hat{C} > 0$, i.e., the importance sampling ratio is bounded. We can then write,
     \begin{align*}
             C_n ~ \expvalue{\bs, \ba \sim \rho^{\pi_e}}{ \norm{\bsigma_n(\bs, \ba)}} & \leq \sum^{d_s}_{i=1} \expvalue{\bs, \ba \sim \rho^{\pi_e}}{C_n \sigma_{n,i}(\bs, \ba)}\\
        &=  \sum^{d_s}_{i=1} \expvalue{\bs, \ba \sim \rho^{\pi_b}}{C_n \sigma_{n,i}(\bs, \ba)\frac{\rho^{\pi_e}(s,a)}{\rho^{\pi_b}(s,a)}} \\
        &\leq \frac{1}{\hat{C}} \sum^{d_s}_{i=1} \expvalue{\bs, \ba \sim \rho^{\pi_b}}{C_n \sigma_{n,i}(\bs, \ba)}.
    \end{align*}
   
    Moreover, by taking a union bound over the dimensions $1, ..., d_s$, Lemma~\ref{lemma:consistency_of_behavioural_policy} implies that with probability greater than $1-\delta$, for any set of i.i.d.~trajectories,
    \begin{equation*}
               \sum^{d_s}_{i=1} \expvalue{\bs, \ba \sim \rho^{\pi_e}}{C_n \sigma_{n,i}(\bs, \ba)}  = \calO\left(d_s\left(1+\sqrt{d_s/\delta}\right)\sqrt{\frac{\gamma^{T+1}_n}{n}}\right).
    \end{equation*}
 Consider a sequence $\{\delta_n\}_{n\geq 0}$ such that $\lim_{n\to 0} \delta_n = 0$, and $\lim_{n\to\infty} d_s\left(1+\sqrt{d_s/\delta_n}\right)\sqrt{\sfrac{\gamma^{T+1}_n}{n}} = 0$ (e.g., $\delta_n = \gamma^{-1}_n$), and let $S_n = \sum^{d_s}_{i=1} C_n \sigma_{n,i}(\bs, \ba)$.
Then we have for all $\epsilon > 0$
\begin{align*}
     \sum^{\infty}_{n=0} \sP\left(S_n > \epsilon \right) &= \sum^{N^*(\epsilon)}_{n=0} \sP\left(S_n > \epsilon \right) + \sum^{\infty}_{n=N^*(\epsilon)} \sP\left(S_n > \epsilon \right) \\ 
     &\leq N^*(\epsilon) +  \sum^{\infty}_{n=N^*(\epsilon)} \delta_n < \infty,
 \end{align*}
 where, $N^*(\epsilon)$ is the smallest integer such that $d_s\left(1+\sqrt{d_s/\delta}\right)\sqrt{\sfrac{\gamma^{T+1}_n}{n}} \leq \epsilon$. This implies that
     \begin{equation*}
        \mathbb{P}\left(\sum^{d_s}_{i=1} \expvalue{\bs, \ba \sim \rho^{\pi_e}}{C_n \sigma^{i}_n(\bs, \ba)} \to 0 \text{ for } n \to \infty \right) = 1.
    \end{equation*}
\end{proof}

\section{HAMBO for Offline Reinforcement Learning}
OPE methods are commonly used in offline reinforcement learning (ORL)~\cite{levine2020tutorial} to recommend/learn an optimal policy. 
Moreover, ORL methods also suffer from distribution shifts and are susceptible to overestimation, i.e., overestimating the performance of the recommended policy. Therefore, in principle, a good COPE method can be applied for ORL applications. 
To this end, we propose a natural modification of \textsc{HAMBO-CA} for ORL.
\begin{equation} \label{eq:pessimistic_value_orl}
\vspaceequation
    \tilde{J}(\pi^*) := \max_{\pi} \min_{\boldeta} J_{\tilde{p}_{\boldeta}}(\pi) ~.
    \vspaceequation
\end{equation}
Our proposed method induces pessimism with respect to the epistemic uncertainty of the learned transition model to tackle distribution shifts. Similar, to \textsc{HAMBO-CA}, we can also use the \textsc{HAMBO-DS1} variant to induce pessimsm.

We compare our \textsc{HAMBO}-based ORL variants to other ORL algorithms 
on the OpenAI Gym tasks from the D4RL benchmark \cite{fu2020d4rl}. Specifically,
we consider the HalfCheetah environment with data sets generated with a random and a medicore-performing policy. Our results are presented in
table~\ref{table:orlresults}. 

The max-min optimization in eq (\ref{eq:pessimistic_value_orl}) is typically very challenging. For our experiments we use the soft actor critic algorithm to train the policy and adversary together (DQN algorithm is used for the \textsc{HAMBO-DA1} variant).

Note, our proposed ORL algorithms recommend the policy with the best lower bound and not the best expected return (see eq~\ref{eq:pessimistic_value_orl}). Therefore, in general, they may fail to recommend the optimal policy. This is the price we pay for inducing robustness in our ORL methods. However, 
in practice (see table~\ref{table:orlresults}) we observer  that the \textsc{HAMBO} based ORL methods perform competitively to the start of the art in the field.

\begin{table}
\resizebox{\linewidth}{!}{
\begin{tabular}{c|c|c|c|c|c|c|c|c}
 & \multicolumn{2}{c|}{Ours} & \multicolumn{4}{c|}{Model-based} & \multicolumn{2}{c}{Model-free} \\
\hline
 & \textsc{HAMBO-CA} & \textsc{HAMBO-DA1} & \cite{rigter2022rambo} & \cite{yu2021combo} & \cite{yu2020mopo} & \cite{Morel} & \cite{kumar2020cql} & \cite{kostrikov2022iql} \\
\hline
HalfCheetah-random & 37.1 & 35.1 & 39.5 & 38.8 & 35.4 & 25.6 & 19.6 & - \\
HalfCheetah-medium & 66.9 & 67.9 & 77.9 & 54.2 & 69.5 & 42.1 & 49.0 & 47.4
\end{tabular}
}
\caption{Comparisons on the HalfCheetah from the D4RL benchmark suite. Results of the other algorithms are taken from \cite{rigter2022rambo}.}
\label{table:orlresults}
\end{table}

\end{document}